%% file: paper.tex
\documentclass[a4]{article}

\input{packages}
\input{macros}
\usepackage[margin=1in]{geometry}

\newif\ifdraft %
\draftfalse
\title{Multi-fidelity Bayesian Optimization \\ with Max-value Entropy Search and its parallelization}

\author[1]{Shion~Takeno}
\author[2]{Hitoshi~Fukuoka}
\author[3]{Yuhki~Tsukada}
\author[4]{Toshiyuki~Koyama}
\author[5]{Motoki~Shiga}
\author[6]{Ichiro~Takeuchi}
\author[7]{Masayuki~Karasuyama}

\affil[1,6,7]{Nagoya Institute of Technology}
\affil[2,3,4]{Nagoya University}
\affil[3,5,7]{Japan Science and Technology Agency}
\affil[5]{Gifu University}
\affil[6,7]{National Institute for Material Science}
\affil[5,6]{RIKEN Center for Advanced Intelligence Project}
\affil[ ]{\textit{takeno.s.mllab.nit@gmail.com, fukuoka.hitoshi@j.mbox.nagoya-u.ac.jp, \{tsukada.yuhki,koyama.toshiyuki\}@material.nagoya-u.ac.jp, shiga\_m@gifu-u.ac.jp, \{takeuchi.ichiro,karasuyama\}@nitech.ac.jp}}

\date{}

\begin{document}
\maketitle



\begin{abstract}
In a standard setting of Bayesian optimization (BO), the objective function evaluation is assumed to be highly expensive.
\emph{Multi-fidelity Bayesian optimization} (MFBO) accelerates BO by incorporating lower fidelity observations available with a lower sampling cost.
In this paper, we focus on the information-based approach, which is a popular and empirically successful approach in BO.
For MFBO, however, existing information-based methods are plagued by difficulty in estimating the information gain.
We propose an approach based on \emph{max-value entropy search} (MES), which greatly facilitates computations by considering the entropy of the optimal function value instead of the optimal input point.
We show that, in our \emph{multi-fidelity MES} (MF-MES), most of additional computations, compared with usual MES, is reduced to analytical computations.
Although an additional numerical integration is necessary for the information across different fidelities, this is only in one dimensional space, which can be performed efficiently and accurately.
Further, we also propose parallelization of MF-MES.
Since there exist a variety of different sampling costs, queries typically occur \emph{asynchronously} in MFBO.
We show that similar simple computations can be derived for asynchronous parallel MFBO.
We demonstrate effectiveness of our approach by using benchmark datasets and a real-world application to materials science data.
\end{abstract}

\section{Introduction}
\label{sec:intro}

\emph{Bayesian optimization} (BO) is a popular machine-learning technique for the black-box optimization problem.
Efficiency of BO has been widely shown in a variety of application areas such as scientific experiments \citep{Wigley2016-Fast}, simulation calculations \citep{Ramprasad2017-Machine}, and tuning of machine-learning methods \citep{Snoek2012-Practical}.
In these scenarios, observing an objective function value is usually quite expensive and thus achieving the optimal value with low querying cost is strongly demanded.

Although standard BO only considers directly querying to an objective function $f(\*x)$, in many practical problems, lower fidelity approximations of the original objective function can be observed.
For example, theoretical computations of physical processes often have multiple levels of approximations by which the trade-off between the computational cost and accuracy can be controlled.
A goal of \emph{multi-fidelity Bayesian optimization} (MFBO) is to accelerate BO by utilizing those lower fidelity observations to reduce the total cost of the optimization.

In this paper, we focus on the information-based approach.
For usual BO without multi-fidelity, which we call \emph{single fidelity BO}, seminal works of this direction are \emph{entropy search} (ES) and \emph{predictive entropy search} (PES) proposed by \citet{Henning2012-Entropy} and \citet{Hernandez2014-Predictive}, respectively.
They define acquisition functions by using information gain for the optimal solution $\*x_* \coloneqq \argmax_{\*x} f(\*x)$.
Unlike classical evaluation measures such as expected improvement, the information-based criterion is a measure of global utility which does not require any additional exploit-explore trade-off parameter.
%
The superior performance of information-based methods have been shown empirically, and then, the same approach has also been extended to the multi-fidelity setting \citep{Swersky2013-Multi,Zhang2017-Information}.

Even in the case of single fidelity BO, however, accurately evaluating information gain is notoriously difficult, which often requires complicated numerical approximations.
For MFBO, evaluating information across multiple fidelities is further difficult.
To overcome this difficulty, we consider a novel information-based approach to MFBO, which is based on a variant of ES called \emph{max-value entropy search} (MES), proposed by \citet{Wang2017-Max}.
MES considers the information gain for $f_* \coloneqq \max_{\*x} f(\*x)$ instead of $\*x_*$.
This greatly facilitates the computation of the information gain because $f_*$ is in one dimensional space unlike $\*x_*$, and they showed superior performance of MES compared with ES/PES.
Our method, called \emph{multi-fidelity MES} (MF-MES), can evaluate the information gain for $f_*$ from an observation of an arbitrary fidelity, and we show that additional expressions, compared with MES, can be derived analytically except for one dimensional integral, which can be calculated accurately and efficiently by using standard numerical integration techniques.
This enables us to obtain more reliable evaluation of information gain easily unlike existing information-based MFBO methods because they contain approximations which are difficult to justify.
%
%
Our MF-MES is also advantageous to other measures of global utility for MFBO, such as the knowledge gradient-based method \citep{Poloczek2017-Multi}, because they are often computationally extremely complicated.
Section~\ref{sec:related-work} discusses related studies in more detail.

Further, we also propose \emph{parallelization} of MF-MES.
Since objective functions have a variety of sampling costs, queries naturally occur \emph{asynchronously} in MFBO.
We extend our information gain so that points currently being queried can be taken into consideration. 
Similarly in the case of MF-MES, we show that a required numerical integration in addition to the sampling of $f_*$ is also reduced to one dimensional space through the integration by substitution.
%
This allows us to obtain the reliable evaluation of the information gain for the parallel extension of MF-MES.


%

Our main contributions are summarized as follows:
\vspace{-1em}
\begin{enumerate}
 \setlength{\itemsep}{0pt}
 \item We develop an information-theoretic efficient MFBO method.
       Na{\"i}ve formulation and implementation of this problem raise computationally challenging issues that need to be addressed by carefully-tuned and time-consuming approximate computations.
       By using several computational tricks mainly inspired by MES~\citep{Wang2017-Max}, we show that this computational bottleneck can be nicely avoided without additional assumptions or approximations.
 \item We develop an information-theoretic asynchronous parallel MFBO method.
       To our knowledge, there are no existing works in this topic --- We believe that our method is useful in many practical experimental design and black-box optimization tasks with multiple information sources with different fidelities and its parallel evaluation.
\end{enumerate}

\vspace{-1em}
We empirically demonstrate effectiveness of our approach by using benchmark functions and a real-world application to materials science data.


\section{Preliminary}
\label{sec:prelim}

In this section, we first briefly review a multi-fidelity extension of \emph{Gaussian process regression} (GPR). 
Suppose that
$y^{(1)}_{\*x}, \ldots, y^{(M)}_{\*x}$
are the observations at $\*x \in \cX \subset \RR^d$ with $M$ different fidelities in which
$y^{(M)}_{\*x}$
is the highest fidelity and
$y^{(1)}_{\*x}$
is the lowest fidelity.
%
%
Each observation is modeled as
$y^{(m)}_{\*x} = f^{(m)}_{\*x} + \epsilon$
in which a random noise
$\epsilon \sim \cN(0,\sigma_{\rm noise}^2)$
is added to the underlying true function $f^{(m)}_{\*x} : \cX \to \RR$.
%
The training data set
$\cD_{n} =\{ (\*x_i, y^{(m_i)}_{\*x_i}, m_i) \}_{i \in [n]}$ contains a set of triplets consisting of an input $\*x_i$, fidelity $m_i \in [M]$,
and an output $y^{(m_i)}_{\*x_i}$, where $[n] := \{1, \ldots, n\}$.

Throughout the paper, we assume that a set of outputs $\{ f_{\*x}^{(m)} \}$ for any set of pairs $(\*x,m)$ are always modeled as the multi-variate normal distribution.
%
Standard multi-output extensions of GPR such as multi-task GPR \citep{Bonilla2008-Multi}, co-kriging \citep{Kennedy2000-Predicting}, and semiparametric latent factor model  (SLFM) \citep{Teh2005-Semiparametric}, satisfy this condition.
We call GPR fitted to observations across multiple fidelities \emph{multi-fidelity Gaussian process regression} (MF-GPR), in general.

MF-GPR defines
a kernel function
$k((\*x_i,m_i),(\*x_j,m_j))$
for a pair of training instances
$(\*x_i,y^{(m_i)}_{\*x_i},m_i)$
and
$(\*x_j,y^{(m_j)}_{\*x_j},m_j)$.
An example of this kernel function in the case of SLFM is shown in appendix~\ref{app:SLFM-model}.
%
By defining a kernel matrix $\*K \in \RR^{n \times n}$ in which the $i,j$ element is defined by
$k((\*x_i,m_i),(\*x_j,m_j))$,
all the fidelities $f^{(1)}, \ldots, f^{(M)}$ are integrated into a GPR model in which predictive mean and variance are
%
$
\mu^{(m)}_{\*x}
=
\*k^{(m)}_n(\*x)^\top
\*C^{-1}
\*y$,
and
$\sigma^{2(m)}_{\*x}
 =
 k((\*x,m),(\*x,m))  
 - \*k^{(m)}_n(\*x)^\top
 \*C^{-1}
 \*k^{(m)}_n(\*x)$,
where
$\*C \coloneqq \*K + \sigma_{\rm noise}^2 \*I$ with the identity matrix $\*I$,
$\*y \coloneqq (y^{(m_1)}_{\*x_1}, \ldots, y^{(m_n)}_{\*x_n})^\top$, and
$\*k^{(m)}_n(\*x) \coloneqq (k((\*x,m),(\*x_1,m_1)), \ldots, k((\*x,m),(\*x_n,m_n)))^\top$.
%
%
For later use, we define
$\sigma^{2(mm^\prime)}_{\*x}$
as the predictive covariance between $(\*x,m)$ and $(\*x,m^\prime)$, i.e., covariance for the identical $\*x$ at different fidelities:
%
$\sigma^{2(mm^\prime)}_{\*x}
= k((\*x,m),(\*x,m^\prime))
  -
 \*k^{(m)}_n(\*x)^\top
 \*C^{-1}
 \*k^{(m^\prime)}_n(\*x)$.

\section{Multi-fidelity Bayesian Optimization with Max-value Entropy}
\label{sec:}


We consider Bayesian optimization (BO) for maximizing the highest fidelity function
$f^{(M)}_{\*x}$
when $M$ different fidelities
$y^{(m)}_{\*x}$
for $m = 1, \ldots, M$ are available to querying. 
The querying cost is assumed to be known as $\lambda^{(m)}$, where $\lambda^{(1)} \leq \lambda^{(2)} \ldots \leq \lambda^{(M)}$.
Our goal is to achieve a higher value with smaller accumulated cost of the queryings.
We call this problem \emph{multi-fidelity Bayesian optimization} (MFBO).
When $M = 1$, MFBO is reduced to the usual black box optimization to which we refer as the single fidelity setting, while we refer to the setting $M \geq 2$ as the multi-fidelity setting.
%

We employ the information-based approach, which has been widely used in the single fidelity BO. 
In particular, our approach is inspired by \emph{max-value entropy search} (MES) proposed by \citet{Wang2017-Max}, which considers \emph{information gain} about the optimal value
$\max_{\*x \in \cX} f(\*x)$
obtained by a querying.
%
In the case of MFBO, we need to consider the information gain for identifying the maximum of the highest fidelity function
$f_* \coloneqq \max_{\*x \in \cX} f^{(M)}_{\*x}$
by observing an arbitrary fidelity observation.
%
We refer to our information-based MFBO as \emph{multi-fidelity MES} (MF-MES).
Although information-based approaches often result in complicated computations, we show that the calculation of our information gain is reduced to simple computations by which stable information evaluation becomes possible.

\subsection{Information Gain for Sequential Querying}
\label{sec:info-gain}

We first consider the case that a query is sequentially issued after the previous one is observed, which we refer to as \emph{sequential querying}.
Suppose that we already have a training data set $\cD_{t}$ and need to determine next $\*x_{t+1}$ and $m_{t+1}$.
We define an acquisition function
\begin{align}
 a(\*x,m)
 \coloneqq
 {
 I (f_* ; f^{(m)}_{\*x} \mid \cD_t)
 }
 ~ / ~
 {\lambda^{(m)}},
 \label{eq:acq}
\end{align}
where $I(f_* ; f^{(m)}_{\*x} \mid \cD_t)$ is the mutual information between $f_*$ and $f^{(m)}_{\*x}$ conditioned on $\cD_t$.
%
By maximizing $a(\*x,m)$, we obtain a pair of the input $\*x$ and the fidelity $m$ which maximally gains information of the optimal value $f_*$ of the highest fidelity per unit cost.

%
The mutual information can be written as the difference of the entropy:
\begin{align}
 \begin{split}
  & I(f_*; f^{(m)}_{\*x} \mid  \cD_t)
  \\
  &
= H(f^{(m)}_{\*x} \mid \cD_t)
  - \EE_{f_* \mid \cD_t}\bigl[ H(f^{(m)}_{\*x} \mid f_*, \cD_t)\bigl],
 \end{split}
 \label{eq:mi_entropy}
\end{align}
where $H(\cdot \mid \cdot)$ is the conditional entropy of $p(\cdot \mid \cdot)$.
The first term in the right hand side can be derived analytically for any fidelity $m$:
$
H(f^{(m)}_{\*x} \mid \cD_t) =
\log\left(
\sigma^{(m)}_{\*x} \sqrt{ 2 \pi e }
\right)
$,
where $e \coloneqq \exp(1)$.
The second term in (\ref{eq:mi_entropy}) takes the expectation over the maximum $f_*$.
Since an analytical formula is not known for this expectation, we employ Monte Carlo estimation by sampling $f_*$ from the current GPR:
\begin{align}
 &
 \EE_{f_* \mid \cD_t}\bigl[ H(f^{(m)}_{\*x} \mid f_*, \cD_t)\bigl]
 \approx
 \sum_{f_* \in \cF_*}
 \frac{ H(f^{(m)}_{\*x} \mid f_*, \cD_t)
 }{ | \cF_* | }
 ,
 \label{eq:EH_y}
\end{align}
where $\cF_*$ is a set of sampled $f_*$.
Note that since this sampling approximation is in one dimensional space, accurate approximation can be expected with a small amount of samples.
In Section~\ref{sec:alg}, we discuss computational procedures of this sampling.
For a given sampled $f_*$, the entropy of
$p(f^{(m)}_{\*x} \mid f_*, \cD_t)$
is needed to calculate in (\ref{eq:EH_y}).
%
To make the computation tractable, we replace this conditional distribution with
$p(f^{(m)}_{\*x} \mid f^{(M)}_{\*x} \leq f_*, \cD_t)$,
i.e., conditioning only on the given $\*x$ rather than requiring $f^{(M)}_{\*x} \leq f_*$ for $\forall \*x \in \cX$.
Note that this simplification has been employed by most of entropy-based BO methods \citep[e.g.,][]{Hernandez2014-Predictive,Wang2017-Max} including MES, and superior performance compared with other approaches has been shown.

For any $\zeta \in \RR$, define
$\gamma^{(m)}_{\zeta}(\*x) \coloneqq (\zeta - \mu^{(m)}_{\*x})/\sigma^{(m)}_{\*x}$
as a function for scaling.
%
When $m = M$, the density function
$p(f^{(m)}_{\*x} \mid f^{(M)}_{\*x} \leq f_*, \cD_t)$
is 
\emph{truncated normal distribution}.
%
%
%
The entropy of truncated normal distribution can be represented as \citep{Michalowicz2014-Handbook}
\begin{align}
 \!
 H(f^{(M)}_{\*x} \mid f^{(M)}_{\*x} \leq f_*, \cD_t)
 \!
 & =
 \!
 \log\rbr{
 \sqrt{2 \pi e } \sigma^{(M)}_{\*x}
 \Phi\bigl(\gamma^{(M)}_{f_*}(\*x)\bigl)
 }
 \nonumber
 \\
 &
 -
 \frac
 {
 \gamma^{(M)}_{f_*}(\*x) \phi\bigl(\gamma^{(M)}_{f_*}(\*x)\bigl)
 }
 { 2 \Phi\bigl(\gamma^{(M)}_{f_*}(\*x)\bigl) },
 \label{eq:H_highf_trunc_highf}
\end{align}
where $\phi$ and $\Phi$ are the probability density function and the cumulative distribution function of the standard normal distribution.

Next, we consider the case of $m \neq M$.
Unlike the case of $m = M$, the density
$p(f^{(m)}_{\*x} \mid f^{(M)}_{\*x} \leq f_*, \cD_t)$
is not the truncated normal.
Since MF-GPR represents all fidelities as one unified GPR, the joint marginal distribution
$p(f^{(M)}_{\*x}, f^{(m)}_{\*x}  \mid \cD_t)$
can be immediately obtained from the two dimensional predictive distribution, from which we obtain
$p(f^{(M)}_{\*x} \mid f^{(m)}_{\*x}, \cD_t )$
as
\begin{align}
 f^{(M)}_{\*x} \mid f^{(m)}_{\*x}, \mathcal{D}_t
 \sim
 \mathcal{N}(u(\*x), s^2(\*x)),
 \label{eq:two-dim-conditional}
\end{align}
where
 $u(\*x)         
 =
 {{\sigma}^{2(mM)}_{\*x} \bigl(f^{(m)}_{\*x}- \mu^{(m)}_{\*x}\bigl)}
 / {{\sigma}^{2(m)}_{\*x}}
 + \mu^{(M)}_{\*x},
 \text{ and }
 s^2(\*x)
 =
 {\sigma^2}^{(M)}_{\*x} -
 {\bigl({\sigma}^{2(mM)}_{\*x} \bigl)^2} /
 {{\sigma}^{2(m)}_{\*x}}$.
By using this conditional distribution, the entropy of $p(f^{(m)}_{\*x} \mid f^{(M)}_{\*x} \leq f_*, \cD_t)$ can be written as follows:
\begin{lem}
 Let
 $Z \coloneqq { 1 } / { \sigma^{(m)}_{\*x}  \Phi(\gamma^{(M)}_{f_*}(\*x)) }$
 and
 $\Psi(f^{(m)}_{\*x}) \coloneqq {\Phi\bigl( (f_* - u(\*x)) / s(\*x) \bigl) \phi\bigl(\gamma^{(m)}_{f^{(m)}_{\*x}}(\*x)\bigl)}$.
 %
 %
 Then, for a given $f_*$, we obtain
 \begin{align}
  \begin{split}
   &
   H(f^{(m)}_{\*x} \mid f^{(M)}_{\*x} \leq f_*, \cD_t)
   \\ &
   =
   - \int 
   Z
   \Psi(f^{(m)}_{\*x})
   \log
   \left(
   Z
   \Psi(f^{(m)}_{\*x})
   \right)
   \mathrm{d} f^{(m)}_{\*x}.
  \end{split}
  \label{eq:H_lowf_trunc_highf}
 \end{align}
 \label{lemma:MFMES_conditional_entropy}
\end{lem}
\vspace{-1em}
See Appendix~\ref{app:proof-lemma1} for the proof.

Lemma~\ref{lemma:MFMES_conditional_entropy} indicates that the entropy is represented through the one dimensional integral over $f_{\*x}^{(m)}$.
%
%
Since the integral is only on the one dimensional space, standard numerical integration techniques (e.g., quadrature) can provide precise approximation efficiently.
Consequently, we see that that the entropy $H(f^{(m)}_{\*x} \mid f_*, \cD_t)$ in \eq{eq:EH_y} can be obtained accurately with simple computations.

\subsection{Asynchronous Parallelization}
\label{sec:parallel}

\begin{figure}[t] 
 \cigr{.6}{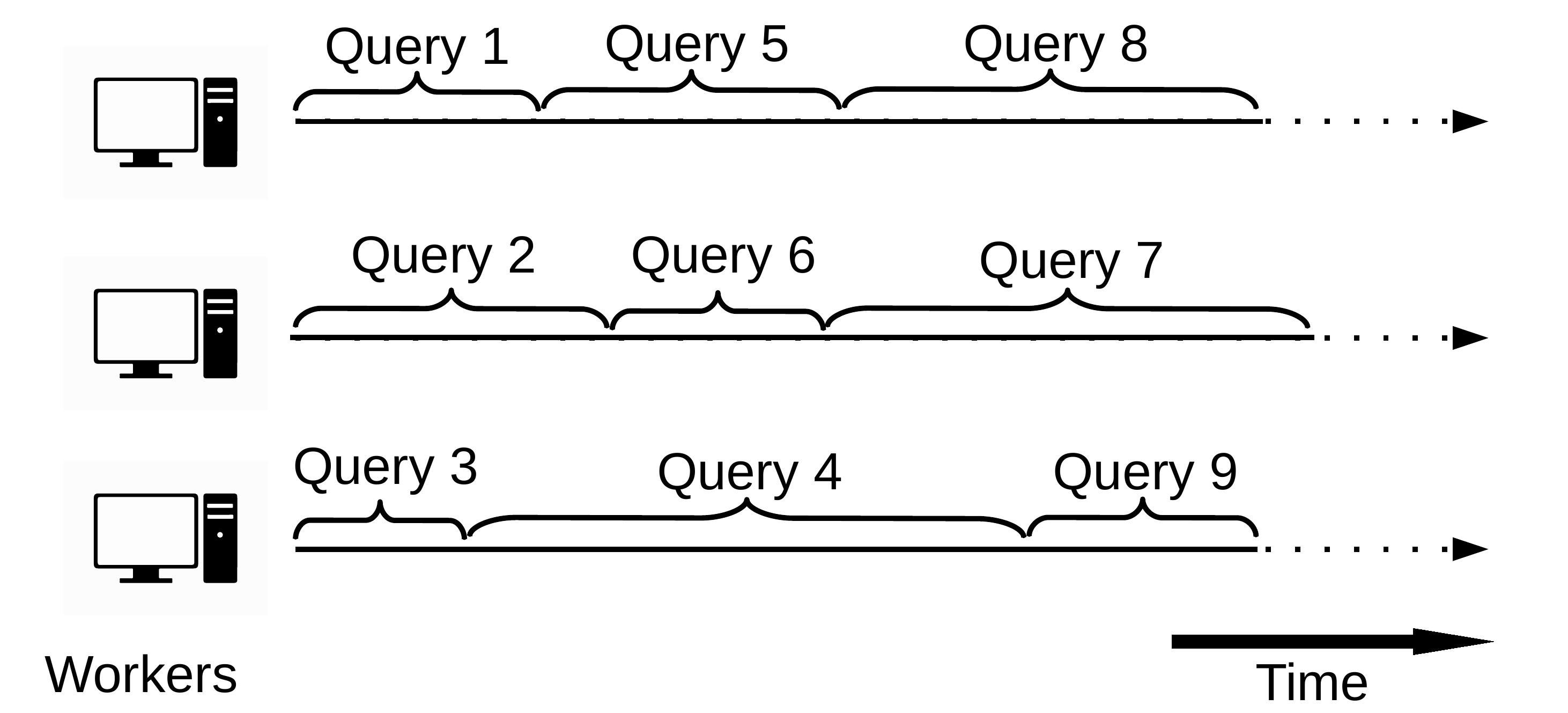}
 \caption{
 Asynchronous parallelization in MFBO.
 Because of diversity of the evaluation cost of objective functions, queries typically occur asynchronously.
 When a worker becomes available, a next query should be determined while taking queries being evaluated in the other workers into consideration.
 }
 \label{fig:asyncBO}
\end{figure}

We consider an extension of MF-MES for the case that multiple queries can be issued in parallel, which we refer to as \emph{parallel querying}.
Suppose that we have $q > 1$ ``workers'' each one of which can evaluate an objective function value.
In the context of parallel BO, the two settings called \emph{synchronous} and \emph{asynchronous} parallelizations can be considered.
As shown in \figurename~\ref{fig:asyncBO}, since MFBO evaluates a variety of different costs of objective functions, queries naturally occur asynchronously.
Thus, we focus on asynchronous parallelization (See Appendix~\ref{app:parallel-sync} for the discussion of the synchronous setting).

\setlength{\textfloatsep}{10pt}

Suppose that
$q - 1$ pairs of the input $\*x$ and the fidelity $m$,
written as
$\cQ \coloneqq \{ (\*x_1, m_1), \ldots, (\*x_{q-1}, m_{q-1}) \}$,
are now being evaluated by using $q - 1$ workers, and an additional query to an available worker needs to be determined.
Let $\*f_{\cQ} \coloneqq (f_{\*x_1}^{(m_1)}, \ldots, f_{\*x_{q-1}}^{(m_{q-1})})^\top$.
%
Then, a natural extension of MF-MES to determine the $q$-th pair $(\*x_q, m_q)$ is
\begin{align}
 a_{\rm para}(\*x, m)
 =
 I(f_* ; f_{\*x}^{(m)} \mid \cD_t, \*f_{\cQ}) ~ / ~ \lambda^{(m)}.
 \label{eq:acq_para}
\end{align}
The numerator is the mutual information conditioned on $\*f_{\cQ}$ which is defined by
\begin{align}
 &
 I(f_* ; f_{\*x}^{(m)} \mid \cD_t, \*f_{\cQ})
 \coloneqq
 \EE_{\*f_{\cQ} \mid \cD_t}
 \left[
 H(f_{\*x}^{(m)} \mid \cD_t, \*f_{\cQ})
 \right]
 \nonumber
 \\ 
 &
 \ \ \
 -
 \EE_{\*f_{\cQ}, f_* \mid \cD_t }
 \left[
 H(f_{\*x}^{(m)} \mid \cD_t, \*f_{\cQ}, f_{\*x}^{(M)} \leq f_*)
 \right].
 \label{eq:minfo_para}
\end{align}
Compared with the mutual information in sequential querying \eq{eq:mi_entropy}, this equation additionally takes the expectation over $\*f_{\cQ}$ which is currently under evaluation.
Thus, by using \eq{eq:acq_para}, we can select a cost effective pair of $\*x$ and $m$ while the $q - 1$ pairs running on the other workers are taken into consideration.

Although \eq{eq:minfo_para} contains the $|\cQ| + 2$ dimensional integral at a glance, we show that this can be calculated by at most $2$ dimensional numerical integral.
Let
$\*\Sigma_{\cM} \in \RR^{2 \times 2}$
and
$\*\Sigma_{\cQ} \in \RR^{q-1 \times q-1}$
be the predictive covariance matrices for
$\cM \coloneqq \{(\*x,m),(\*x,M)\}$
and $\cQ$, respectively, and
$\*\Sigma_{\cQ,\cM} (= \*\Sigma_{\cM,\cQ}^\top) \in \RR^{q-1 \times 2}$
be the predictive covariance matrix of the rows $\cQ$ and the columns $\cM$.
For later use, we define the conditional distribution
$p(f_{\*x}^{(m)}, f^{(M)}_{\*x} \mid \cD_t, \*f_{\cQ})$
as follows
\[
 \begin{bmatrix}
  f_{\*x}^{(m)} \\
  f_{\*x}^{(M)}
 \end{bmatrix}
 \mid
 \cD_t,
 \*f_{\cQ}
 \sim
 \cN
 \left(
 \begin{bmatrix}
  \mu_{\*x \mid \*f_{\cQ}}^{(m)} \\
  \mu_{\*x \mid \*f_{\cQ}}^{(M)}
 \end{bmatrix},
 \begin{bmatrix}
  \sigma^{2(m)}_{\*x \mid \*f_{\cQ}} & \sigma^{2(mM)}_{\*x \mid \*f_{\cQ}}
  \\
  \sigma^{2(mM)}_{\*x \mid \*f_{\cQ}} & \sigma^{2(M)}_{\*x \mid \*f_{\cQ}}
 \end{bmatrix}
 \right),
\]
where
\begin{align}
 &
 \begin{bmatrix}
  \mu_{\*x \mid \*f_{\cQ}}^{(m)} \\
  \mu_{\*x \mid \*f_{\cQ}}^{(M)}
 \end{bmatrix}
 =
 \begin{bmatrix}
  \mu_{\*x}^{(m)} \\
  \mu_{\*x}^{(M)}
 \end{bmatrix}
 +
 \*\Sigma_{\cM,\cQ}
 \*\Sigma_{\cQ}^{-1}
 (\*f_{\cQ} - \*\mu_{\cQ}),
 \label{eq:mu_given_Q}
 \\ 
 &
 \begin{bmatrix}
  \sigma^{2(m)}_{\*x \mid \*f_{\cQ}} & \sigma^{2(mM)}_{\*x \mid \*f_{\cQ}}
  \\
  \sigma^{2(mM)}_{\*x \mid \*f_{\cQ}} & \sigma^{2(M)}_{\*x \mid \*f_{\cQ}}
 \end{bmatrix}
 =
 \*\Sigma_{\cM} -
 \*\Sigma_{\cM,\cQ}
 \*\Sigma_{\cQ}^{-1}
 \*\Sigma_{\cQ,\cM},
 \label{eq:sigma_given_Q}
\end{align}
and
$\*\mu_{\cQ} \coloneqq (\mu_{\*x_1}^{(m_1)}, \ldots, \mu_{\*x_{q-1}}^{(m_{q-1})})^\top$.
Note that \eq{eq:mu_given_Q} is a random variable vector because it depends on $\*f_{\cQ}$, while all the elements of \eq{eq:sigma_given_Q} are constants.
By using these equations, the mutual information \eq{eq:minfo_para} is re-written as follows:
\begin{lem}
 Let
 \begin{align}
  \tilde{f}_*
  & \coloneqq
  f_* - \mu_{\*x \mid \*f_{\cQ}}^{(M)},
  \label{eq:tilde_f_star}
 \end{align}
 and
 $\tilde{f}_{\*x}^{(m)}
 \coloneqq
 f_{\*x}^{(m)} - \mu_{\*x \mid \*f_{\cQ}}^{(m)}$.
 %
 Then, we obtain
  \begin{align}
    \begin{split}
     & I(f_* ; f_{\*x}^{(m)} \mid \cD_t, \*f_{\cQ}) =
     \log
     \left(
     \sigma_{\*x \mid \*f_{\cQ}}^{(m)} \sqrt{2 \pi e}
     \right)
     \\
     & \ 
     -
     \EE_{\tilde{f}_* \mid \cD_t}
     \left[
     \int
     \eta(\tilde{f}_*, \tilde{f}_{\*x}^{(m)})
     \log \eta(\tilde{f}_*, \tilde{f}_{\*x}^{(m)})
     ~ {\rm d} \tilde{f}_{\*x}^{(m)}
     \right]
    \end{split}
    \label{eq:minfo_para_fast}
   \end{align}
   where
\begin{align}
 \begin{split}
  \eta(\tilde{f}_*, \tilde{f}_{\*x}^{(m)})
  &
  \! \!
  \coloneqq
  \! \!
  \frac{
  \Phi\left(
   \frac{
   \tilde{f}_* - \left( \sigma_{\*x \mid \*f_{\cQ}}^{2 (mM)} ~ / ~ \sigma_{\*x \mid \*f_{\cQ}}^{2 (m)} \right) \tilde{f}^{(m)}_{\*x}
   }
   {   
   \sigma_{\*x \mid \*f_{\cQ}}^{2 (M)}
   - \left( \sigma_{\*x \mid \*f_{\cQ}}^{2 (mM)} \right)^2 ~ / ~ \sigma_{\*x \mid \*f_{\cQ}}^{2 (m)}
   }
  \right)
  \phi\left(
  \frac{ \tilde{f}^{(m)}_{\*x} }{ \sigma^{(m)}_{\*x \mid \*f_{\cQ}}}
  \right)
  }
  {
  \sigma^{m}_{\*x \mid \*f_{\cQ}}
  \Phi\left( \frac{ \tilde{f}_* }{ \sigma^{(M)}_{\*x \mid \*f_{\cQ}} } \right)
  }.
 \end{split}
 \label{eq:eta}
\end{align}
\label{lemma:paraMFMES_minfo}
\end{lem}
\vspace{-1em}
See Appendix~\ref{app:proof-para} for the proof.
It should be noted that the second term of \eq{eq:minfo_para_fast} only contains the integral over two variables ($\tilde{f}_{\*x}^{(m)}$ and $\tilde{f}_{*}$) unlike the original formulation \eq{eq:minfo_para}.
The first term of \eq{eq:minfo_para_fast} can be directly calculated because
$\sigma_{\*x \mid \*f_{\cQ}}^{(m)}$
does not depend on the random vector
$\*f_{\cQ}$
as shown in \eq{eq:sigma_given_Q}.
We calculate the expectation in the second term of \eq{eq:minfo_para_fast} by using the Monte Carlo estimation with sampled $\tilde{f}_*$:
\begin{align}
 \sum_{\tilde{f}_* \in \widetilde{\cF}_*}
 \frac{1}{|\widetilde{\cF}_*|}
 \int
 \eta(\tilde{f}_*, \tilde{f}_{\*x}^{(m)})
 \log \eta(\tilde{f}_*, \tilde{f}_{\*x}^{(m)})
 ~ {\rm d} \tilde{f}_{\*x}^{(m)}
 \label{eq:H1_para}
\end{align}
where $\widetilde{\cF}_*$ is a set of sampled $\tilde{f}_*$.
The integral in this equation can be easily evaluated by using quadrature because it is on the one dimensional space and $\eta(\tilde{f}_*, \tilde{f}_{\*x}^{(m)})$ can be analytically calculated from the definition \eq{eq:eta}.
Further, when $m = M$, this integral is also can be analytically calculated (See Appendix~\ref{app:analytic-entropy-para}).

\section{Computations} 
\label{sec:alg}

Algorithm~\ref{alg:MF-MES} shows the procedure of MF-MES for sequential querying.
%
As the first step in the every iteration, a set of max values $\cF_*$ are sampled from $p(f_* \mid \cD_t)$.
There are several approaches to sampling the max value.
\citet{Wang2017-Max} showed that the effective approximation is possible by using sampling through Gumbel distribution or random feature map (RFM).
Gumbel distribution is widely known in extreme value theory \citep{Gumbel1958-Statistics} as one of generalized extreme value distributions.
%

%
Although the Gumbel approximation is performed under an independent approximation of GPR, \citet{Wang2017-Max} showed the accurate approximation can be obtained.
%
In contrast, RFM \citep{Rahimi2008-Random} can incorporate dependency in the GPR model by using a set of pre-defined basis functions $\*\phi(\*x, m) \in \RR^D$, and the highest fidelity function is represented as
$f^{(M)}_{\*x} \approx \*w^\top \*\phi(\*x, M)$,
where
$\*w \in \RR^D$
(Appendix~\ref{app:SLFM-RFM} shows an example of an RFM approximation in the case of SLFM).
The max value is sampled by maximizing
$\*w^\top \*\phi(\*x, M)$
with respect to $\*x$.
%
For further detail of these two approaches, see \citep{Wang2017-Max}, in which it is also shown that MES is empirically robust with respect to this sampling, and theoretically, they showed that the regret bound can be guaranteed even only for one sample of $f_*$.

\setlength{\textfloatsep}{10pt}
\begin{algorithm}[t]
 \caption{MF-MES for sequential querying}
 \label{alg:MF-MES}
 \begin{algorithmic}[1]
  \Function{MF-MES}{$\cD_0, M, \cX, \{ \lambda^{(m)} \}_{m=1}^M$}
  \For{$t = 0, \ldots, T$}
  \State Generate $\cF_*$ from current $f^{(M)}(\*x)$
  \State $(\*x_{t+1}, m_{t+1}) \leftarrow \argmax_{\*x \in \cX, m}$
  \par  \hskip2em
  {\textproc{InfoGain}}($\*x$, $m$, $\cF_*$, $\cD_t$) $/$ $\lambda^{(m)}$
  \State $\cD_{t+1} \leftarrow \cD_{t} \cup (\*x_{t+1}, y^{(m_{t+1})}(\*x_{t+1}), m_{t+1})$
  \EndFor
  \EndFunction
  \Function{InfoGain}{$\*x$, $m$, $\cF_*$, $\cD_t$}
  \State Calculate $\mu^{(m)}_{\*x}$ and $\sigma^{(m)}_{\*x}$
  \State Set $H_0 \leftarrow \log\left( \sigma^{(m)}_{\*x} \sqrt{ 2 \pi e } \right)$
  \If{$m = M$}
  \State Set $H_1 \leftarrow \sum_{f_* \in \cF_*} \frac{ H(f^{(M)}_{\*x} \mid  f^{(M)}_{\*x} \leq f_*, \cD_t) }{|\cF_*|}$   
  \par \hskip2em by using \eq{eq:H_highf_trunc_highf}
  \Else
  \State Calculate $\mu^{(M)}_{\*x}$ and $\sigma^{(M)}_{\*x}$ and $\sigma^{2(mM)}_{\*x}$
  \State Set $H_1 \leftarrow \sum_{f_* \in \cF_*} \frac{H(f^{(m)}_{\*x} \mid f^{(M)}_{\*x} \leq f_*, \cD_t) }{|\cF_*|}$
  \par \hskip2em by using \eq{eq:H_lowf_trunc_highf}
  \EndIf
   \State Return $H_0 - H_1$
  \EndFunction
 \end{algorithmic}
\end{algorithm}


Once $\cF_*$ is generated, the acquisition function calculation can be analytically performed except for one dimensional numerical integration.
Although most complicated process in the algorithm is the calculation of \eq{eq:H_lowf_trunc_highf} shown in line 15 of Algoirthm~\ref{alg:MF-MES}, this is also quite simple in practice as described below.
For a given $f_*$ and the conditional distribution \eq{eq:two-dim-conditional} which is constructed from the two dimension GPR predictive distribution $p(f^{(M)}_{\*x}, f^{(m)}_{\*x}  \mid \*x, \cD_t)$,
the integral of \eq{eq:H_lowf_trunc_highf} can be computed by $O(1)$.
%
Further, since
\eq{eq:two-dim-conditional}
does not depend on sampled $f_*$, it is not required to re-calculate
\eq{eq:two-dim-conditional}
for each one of sampled $f_*$.

For the acquisition function maximization ($\argmax$ in line 4), if the candidate space $\cX$ is a discrete set, we simply calculate the acquisition values for all $\*x \in \cX$.
For a continuous space, popular approaches such as DIRECT \citep{Jones1993-Lipschitzian}
and gradient-based optimizers are applicable.
Note that our acquisition function is differentiable, and the derivative of the integral (\ref{eq:H_lowf_trunc_highf}) can be calculated by the same one dimensional numerical integral procedure.

For the case of parallel querying, the acquisition function maximization is performed when a worker becomes available.
To evaluate \eq{eq:H1_para}, we need to sample $\tilde{f}_*$, which is determined through $f_*$ and $\*f_{\cQ}$ as shown in \eq{eq:tilde_f_star}. 
This can be easily performed through RFM.
By calculating
$\*w^\top \*\phi(\*x, m)$
for $(\*x, m) \in \cQ$ with the sampled parameter $\*w$, we can directly obtain a sample of $\*f_{\cQ}$.
For $f_*$, we maximize $\*w^\top \*\phi(\*x, M)$ as in the sequential querying case.
The algorithm of Parallel MF-MES is shown in Appendix~\ref{app:alg-para}.
%

Throughout the paper, we use $I(f_* ; f_{\*x}^{(m)})$ as the information gain for brevity.
$I(f_* ; y^{(m)}_{\*x})$, in which noisy observation $y^{(m)}_{\*x}$ is contained, is also possible to use with the almost same procedure (for details, see Appendix~\ref{app:info-gain-with-noise}).

Although we mainly focus on the case that we only have the discrete fidelity level $m \in \{ 1, \ldots, M \}$ as an ``ordinal scale'', several studies consider the setting in which a fidelity can be defined as a point $z$ in a continuous ``fidelity feature'' (FF) space $\cZ$ \citep{Kandasamy2017-Multi}.
This setting is more restrictive because it requires additional side-information $z$ which specifies a degree of fidelity, though this prior knowledge may be able to improve the accuracy.
By introducing a kernel function in fidelity space $\cZ$, our method can easily adapt to this setting (See appendix~\ref{app:extension-z-space}).
\section{Related Work}
\label{sec:related-work}

Multi-fidelity extension of BO has been widely studied.
For example, \citep{Huang2005-Sequential, Lam2015-Multifidelity, Picheny2013-Quantile} extended the standard EI to the multi-fidelity setting.
As with the usual EI, these are local measures of utility unlike the information-based approaches.
\emph{Gaussian process upper confidence bound} (GP-UCB) \citep{Srinivas2010-Gaussian} is a popular approach in the single fidelity setting, and some studies proposed its multi-fidelity extensions.
\citet{Kandasamy2016-Gaussian} proposed multi-fidelity GP-UCB for discrete fidelity $m = 1, \ldots, M$, and further, \citet{Kandasamy2017-Multi} proposed a similar UCB-based approach for the setting with the continuous fidelity space $\cZ$.
However, the UCB criterion has a trade-off parameter which balances exploit-exploration.
In practice, this parameter needs to be carefully selected to achieve good performance.
%
%
Another approach recently proposed in \citep{Sen2018-Multi} is a multi-fidelity extension of a hierarchical space partitioning \citep{Bubeck2011-X}.
However, this method assumes that the approximation error can be represented as a known function form of cost, and
%
%
further, they associate fidelity with the depth of hierarchical tree, but the appropriateness of a specific choice of a pair of a point $\*x$ and fidelity $m$ is difficult to interpret.

%
Information-based BO has also been studied for the multi-fidelity setting, including \emph{entropy search} (ES)-based \citep{Swersky2013-Multi,Klein2017-Fast} and \emph{predictive entropy search} (PES)-based \citep{Zhang2017-Information,McLeod2018-Practical} methods.
Although these methods can measure global utility of the query without introducing any trade-off parameter, they inherit the computational difficulty of the original ES and PES, which consider the entropy of $p(\*x_*)$, where $\*x_* \coloneqq \argmax_{\*x} f(\*x)$ is the optimal solution.
%
%
PES mitigates computational difficulty by using 1) the symmetric property of the mutual information, and 2) several assumptions which simplify involved densities.
However, integral with respect to $\*x_*$ is still necessary though the dimension of $\*x_*$ can be high, and the complicated approximation procedure including \emph{expectation propagation} \citep{Minka2001-Minka} is required.
%
Further, an additional assumption about inter-fidelity differences are required in the case of \citep{Zhang2017-Information}.
%
\citet{Song2018-General} proposed another information-based approach, which separates phases of the low-fidelity exploration and the highest fidelity optimization.
%
%
%
However, the transition of these phases are controlled by a hyper-parameter which is necessary to set appropriately beforehand.

Another approach incorporating a measure of global utility is \emph{knowledge gradient} (KG)-based methods \citep{Poloczek2017-Multi,Wu2017-Continuous}.
This approach evaluates the max gain of predictive mean $\max_{\*x \in \cX} \mu^{(M)}_{\*x}$.
In particular, misoKG \citep{Poloczek2017-Multi} deals with the discrete fidelity case.
However, the acquisition function evaluation
requires the expected value of the maximum of the mean function $\EE [ \max_{\*x' \in \cX} \mu^{(M)}_{\*x'} ]$ after adding $y^{(m)}_{\*x}$ into training set, meaning that the maximization of the acquisition function is defined as a nested optimization.
%
%
%
Although a variety of computational techniques have been studied for KG, this nested optimization process is highly cumbersome to implement and computationally expensive. 

In contrast, our MF-MES is based on much simpler computations compared with existing information-based methods and other measures of global utility.
Original MES calculates the entropy by representing a conditional distribution of $f_{\*x}$ given $f_*$ as a truncated normal distribution.
As we saw in Section~\ref{sec:info-gain}, for the information gain from a lower fidelity, the truncated normal approach is not applicable anymore because lower fidelity functions $f^{(m)}_{\*x}$ for $m = 1, \ldots, M - 1$ are not truncated for a given $f_*$.
We already show that equations derived in Lemma~\ref{lemma:MFMES_conditional_entropy} enables us to evaluate the entropy accurately with the only one dimensional additional numerical integration.
%
%
For further acceleration of MES, \citet{Ru18-Fast} proposed approximating the density of $f_*$ and $f$ given $f_*$  by normal distributions, but reliability of these approximations are not clearly understood, and thus we do not employ in this paper.

The parallel extension of BO has been widely studied \citep[e.g.,][]{Snoek2012-Practical,Desautels2014-Parallelizing}.
As we described in Section~\ref{sec:parallel}, MFBO is typically asynchronous, while many of existing studies focus on the synchronous setting including PES-based parallel BO \citep{Amar2015-Parallel}.
Several papers focus on the asynchronous setting \citep{Kandasamy2018-Parallelised}, but these methods are difficult to apply to the multi-fidelity setting because they do not provide any criterion to select fidelity.
To our knowledge, an extension of KG \citep{Wu2017-Continuous} is an only parallel method proposed for MFBO.
However, this method is only for the synchronous setting, and further, it is only shown for the FF-based setting which is more restrictive as we described in the end of Section~\ref{sec:alg}.
%
%
We also note that a parallel extension of MES has not been shown even for the single-fidelity setting.
About a possible sequential/parallel settings of MF-MES, a summary is shown in Appendix~\ref{app:summary-of-settings}.

\section{Experiments}
\label{sec:exp}

We evaluate effectiveness of MF-MES compared with other existing methods.
To evaluate performance, we employed simple regret (SR) and inference regret (IR).
SR is defined by
$\max_{\*x \in \cX} f^{(M)}(\*x) - \linebreak \max_{i \in \{i \mid i \in [t], m_i = M\}} f^{(M)}(\*x_i)$,
indicating the error by the best point queried so far.
IR is defined by
\linebreak $\max_{\*x \in \cX} f^{(M)}(\*x) - f^{(M)}(\hat{\*x}_t)$,
where
$\hat{\*x}_t \coloneqq \argmax_{\*x \in \cX} \mu^{(M)}_{\*x}$
which is seen as the recommendation from the model at iteration $t$.
If IR is larger than SR at an iteration, we employed the value of SR as IR of that iteration for stable evaluation.
%
%
For MF-GPR, we used SLFM in GP-based methods, unless otherwise noted.
For the kernel function, we used Gaussian kernel with automatic relevance determination (ARD).

We used a synthetic function generated by GPR, two benchmark functions, and a real-world dataset from materials science.
For the GP-based synthetic function, we generated $d = 3$ dimensional synthetic functions through an SLFM model which has two fidelity levels.
The benchmark functions are called Styblinski-Tang, and HartMann6, which has $M = 2, \text{ and } 3$ fidelities, respectively.
The sampling cost is set $(\lambda^{(1)}, \lambda^{(2)}) = (1, 5)$ when $M = 2$, and $(\lambda^{(1)}, \lambda^{(2)}, \lambda^{(3)}) = (1, 3, 5)$ when $M = 3$.
As an example of practical applications, we applied our method to the parameter optimization of a simulation model in materials science.
The task is to optimize two material parameters of the model \citep{Tsukada2014-Equilibrium} by minimizing the discrepancy between the precipitate shape predicted by the model and one measured by an electron microscope.
%
The relative cost of the objective function evaluation is determined by the accuracy of the computational model which is specified beforehand as
$(\lambda^{(1)}, \lambda^{(2)}, \lambda^{(3)}) = (5, 10, 60)$.
Unlike benchmark functions, the candidate $\*x$ is fixed beforehand in this dataset (so-called the pooled setting).
Each fidelity has 62,500 candidate points.
The experiments on the GP-based synthetic function were performed 100 times (10 different initialization for each one of 10 generated functions).
The other benchmark functions and the material dataset were performed 10 times with different initialization.
For further detail of the settings, see Appendix~\ref{app:settings}.

\begin{figure*}
\subfloat[Simple regret.]{
\igr{.24}{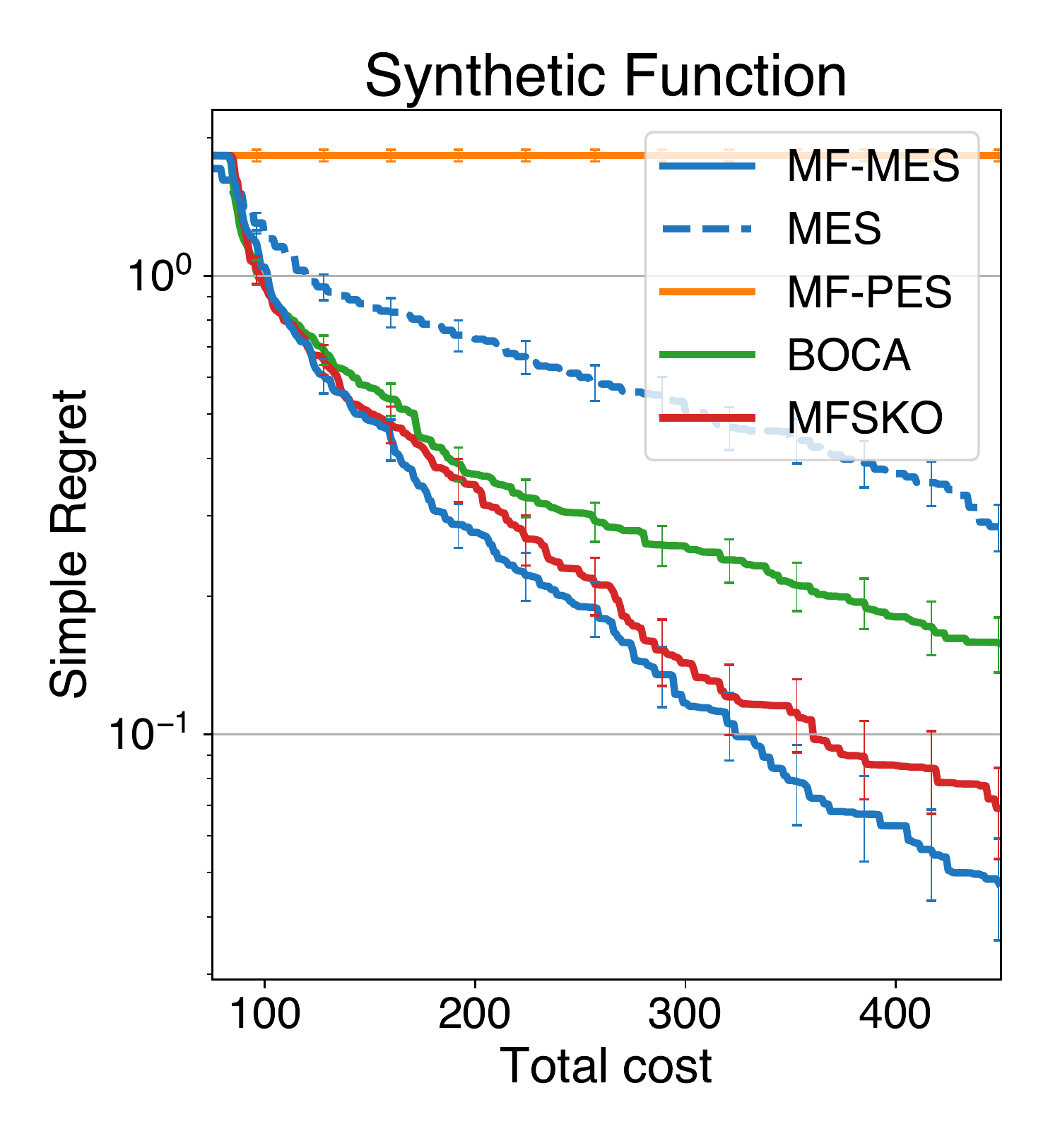}
\igr{.24}{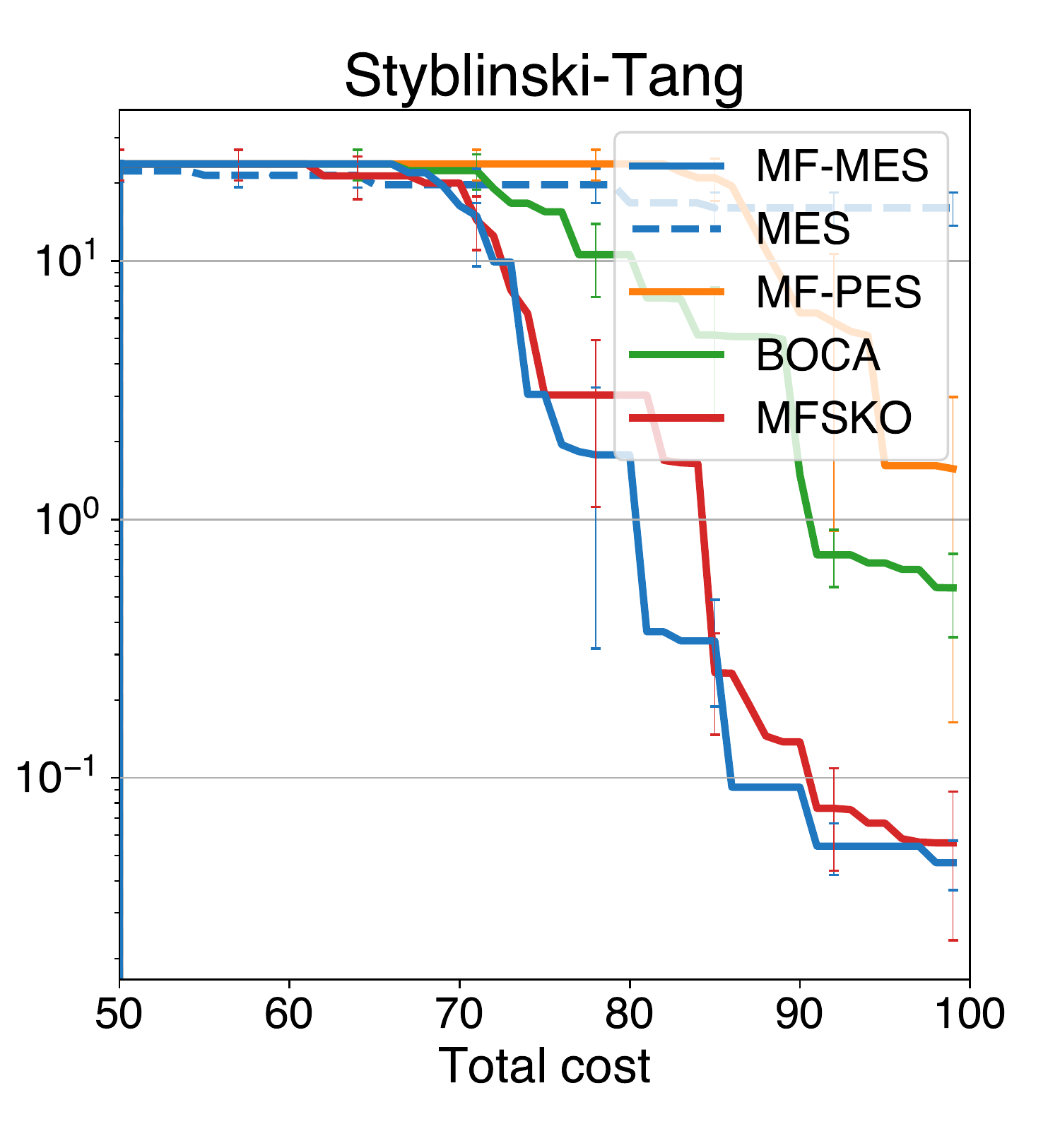}
\igr{.24}{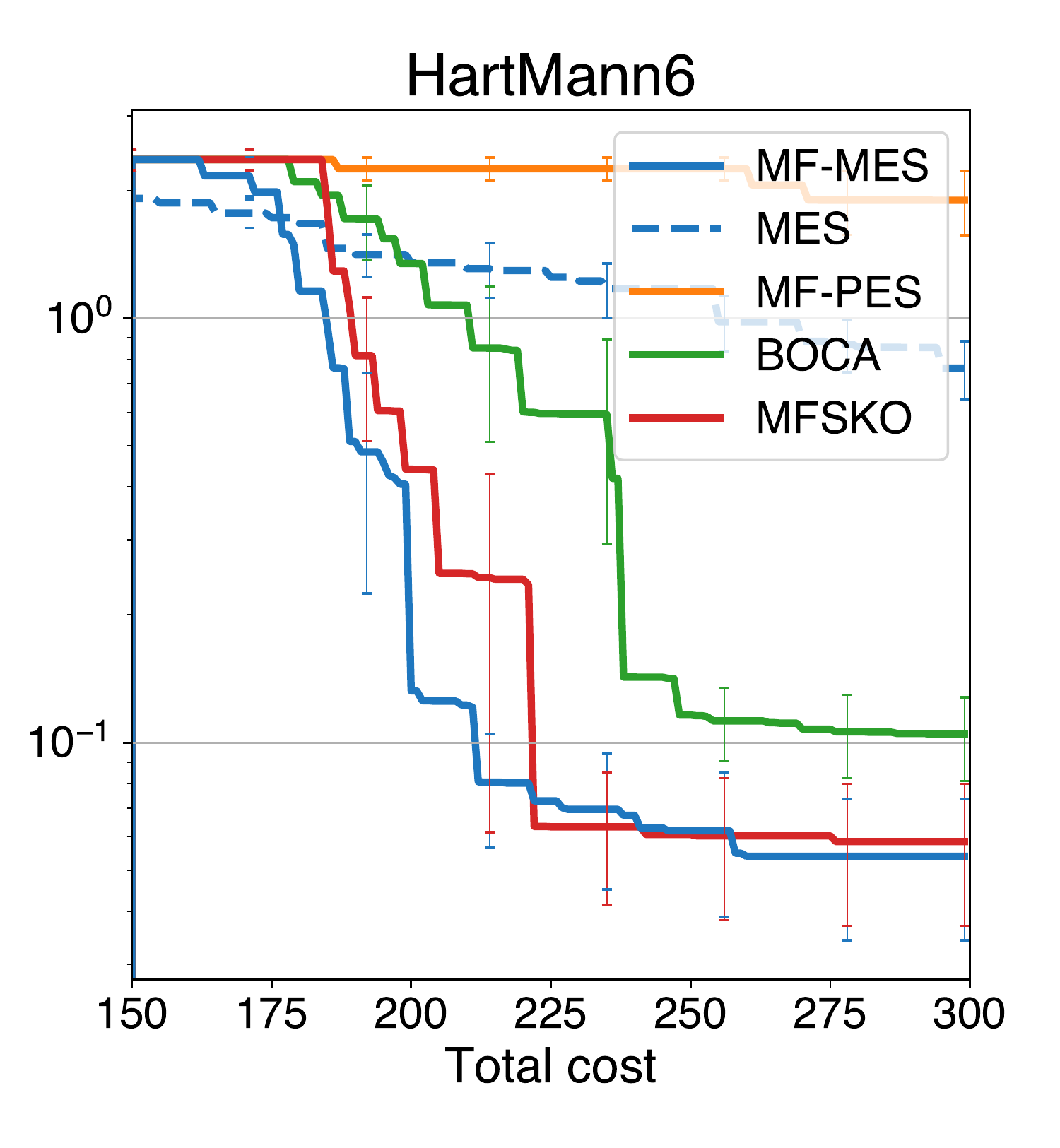}
\igr{.24}{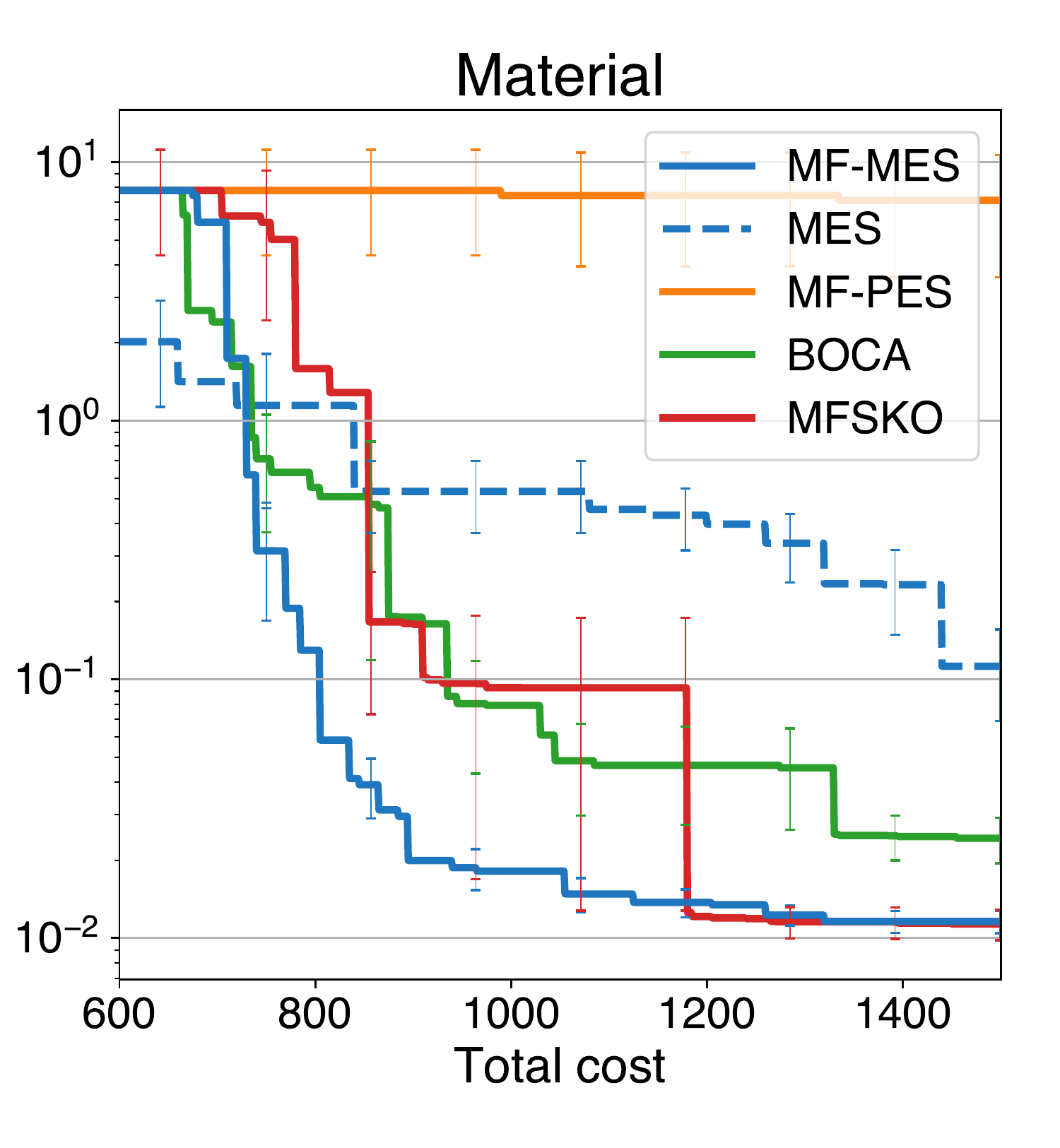}
}

\subfloat[Inference regret.]{
\igr{.24}{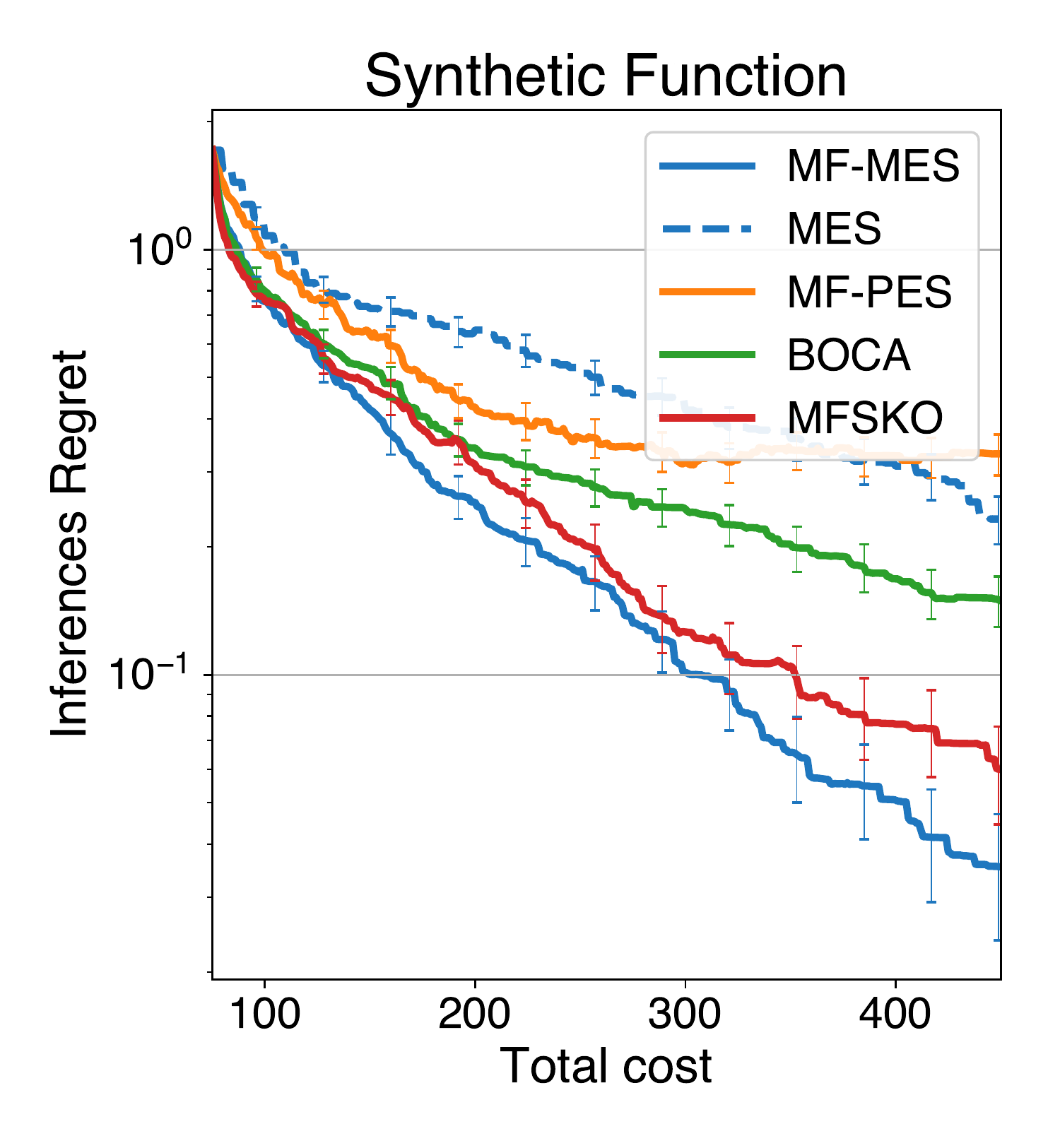}
\igr{.24}{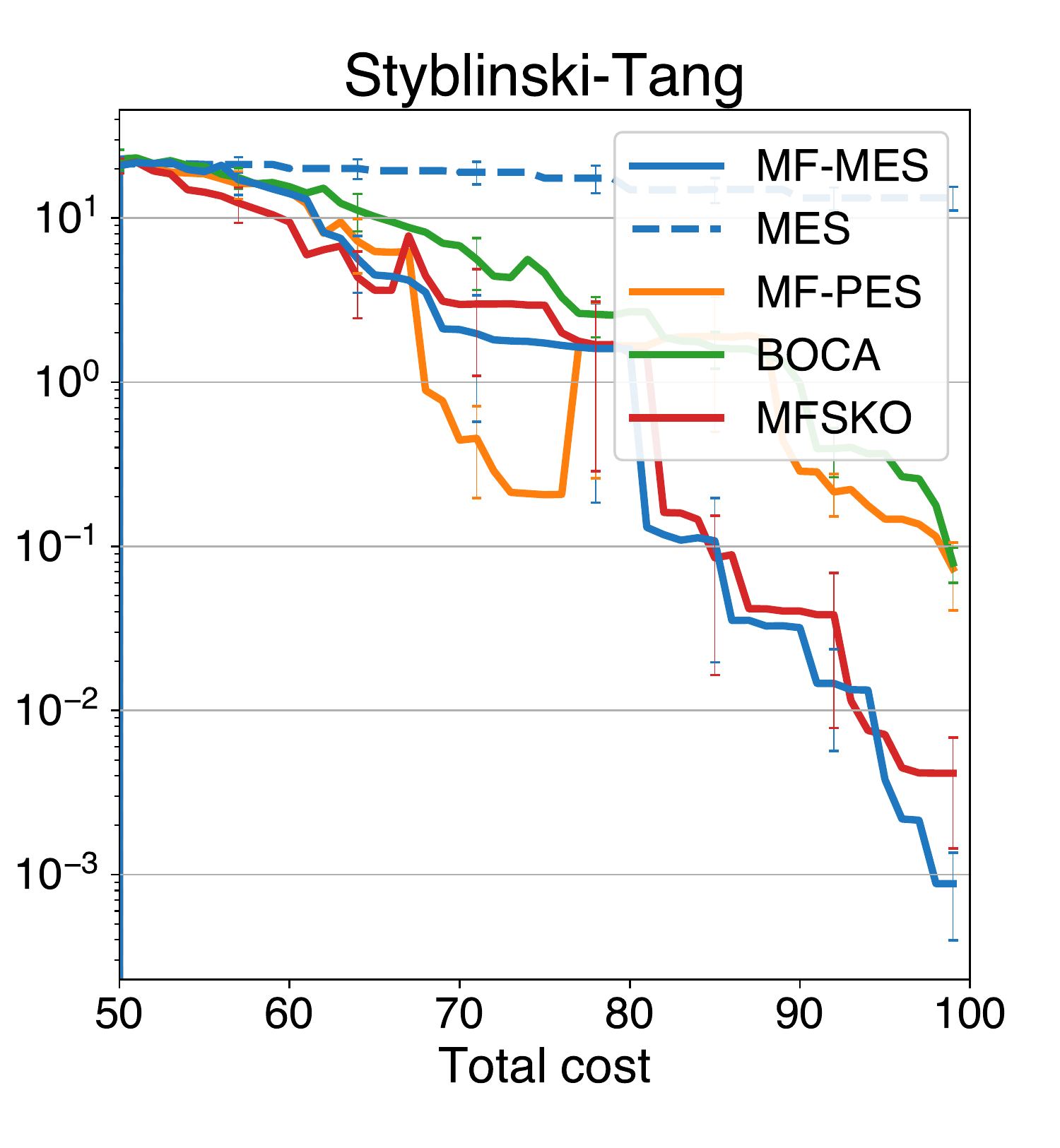}
\igr{.24}{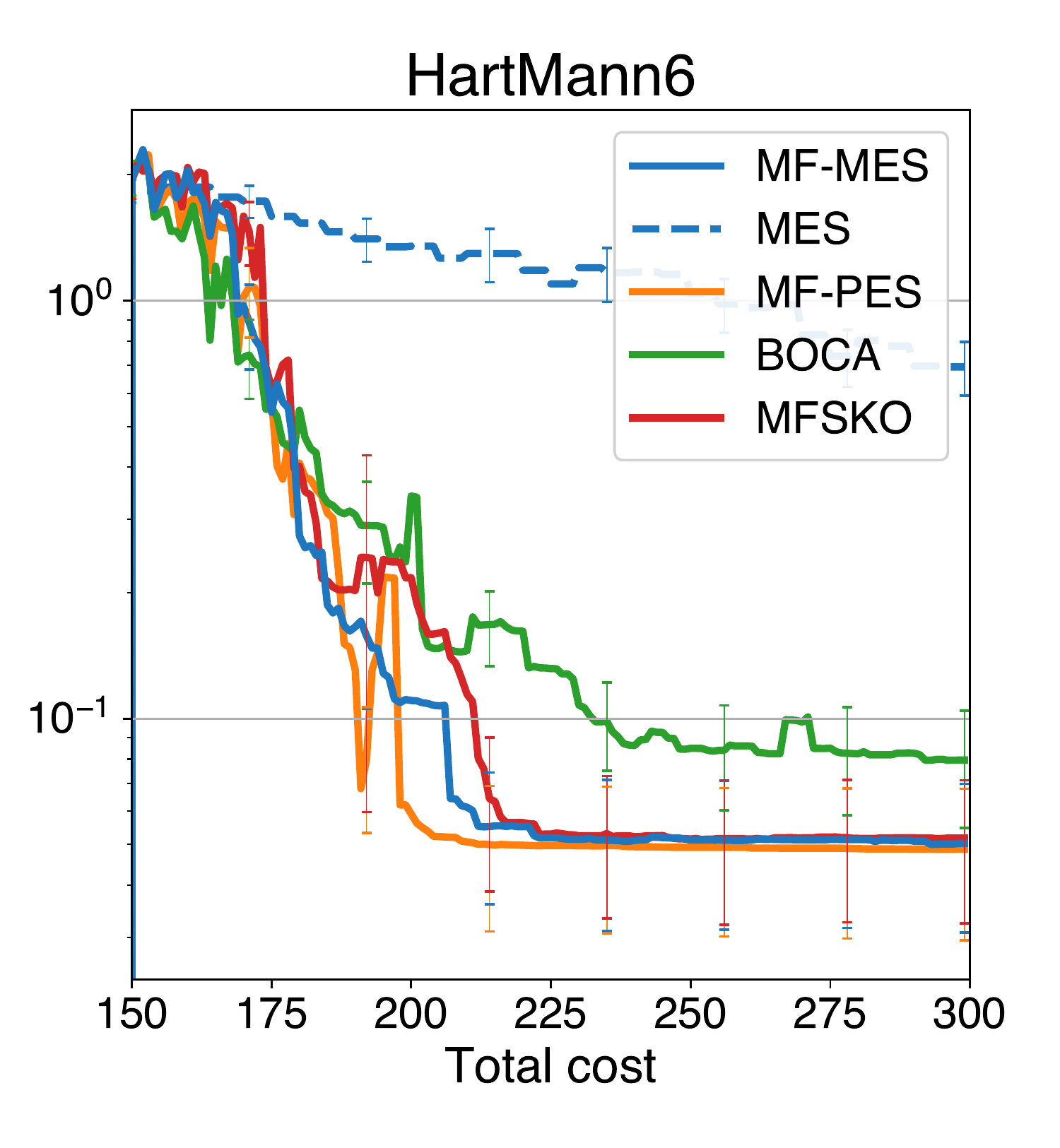}
\igr{.24}{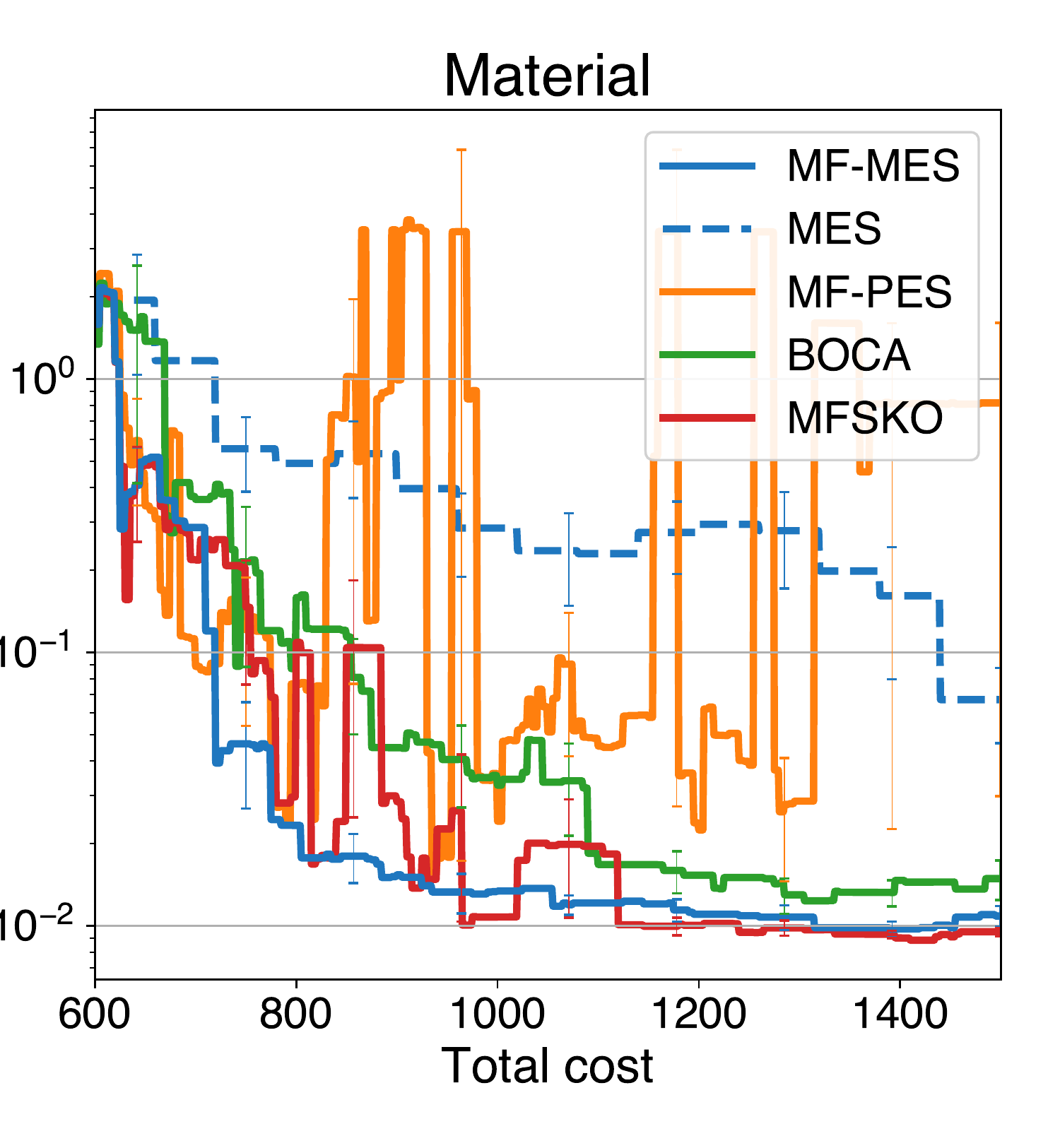}
}
\caption{Performance comparison on sequential querying.}
\label{fig:regret-sequential}
\subfloat[Simple regret.]{
\igr{.24}{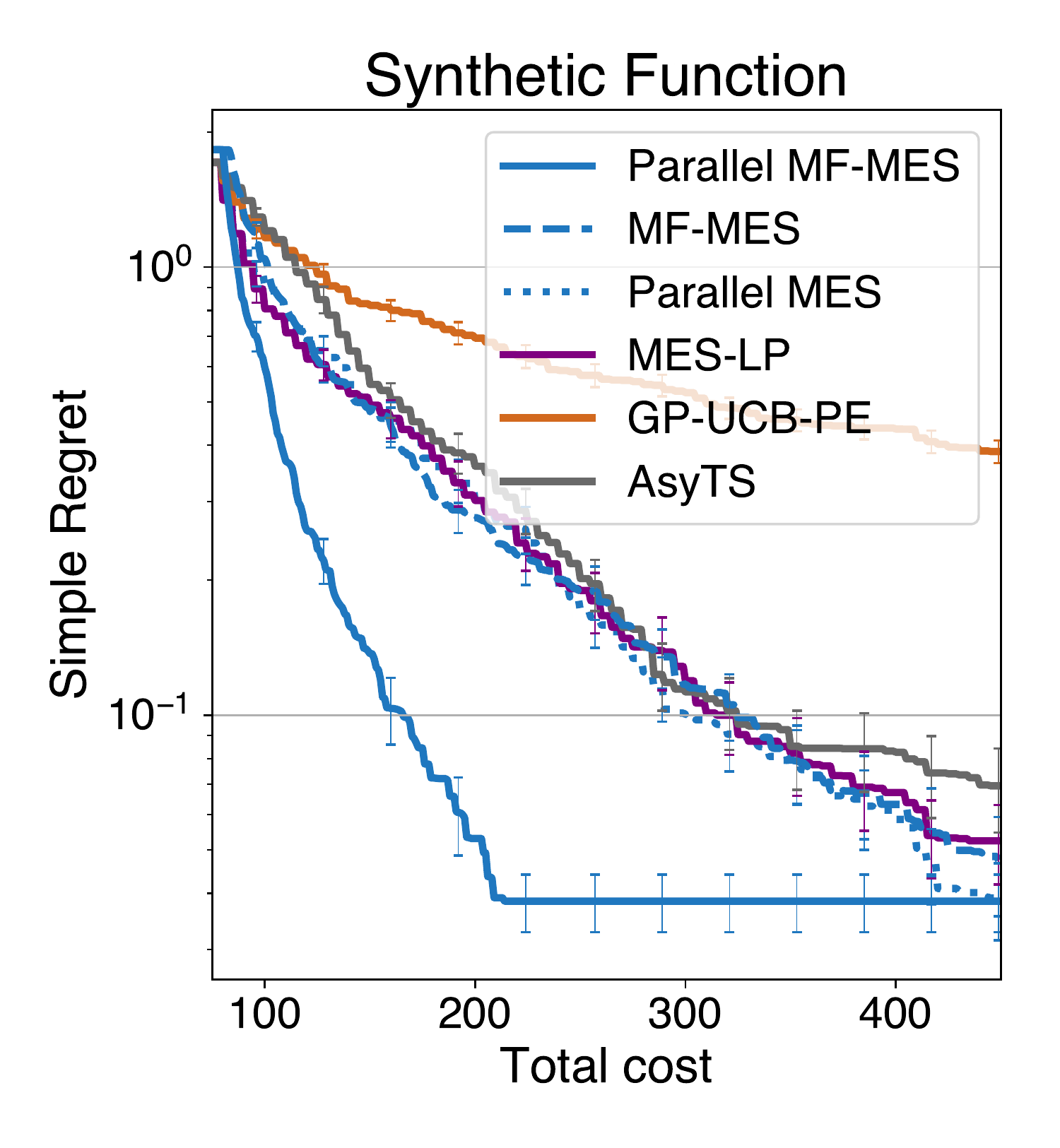}
\igr{.24}{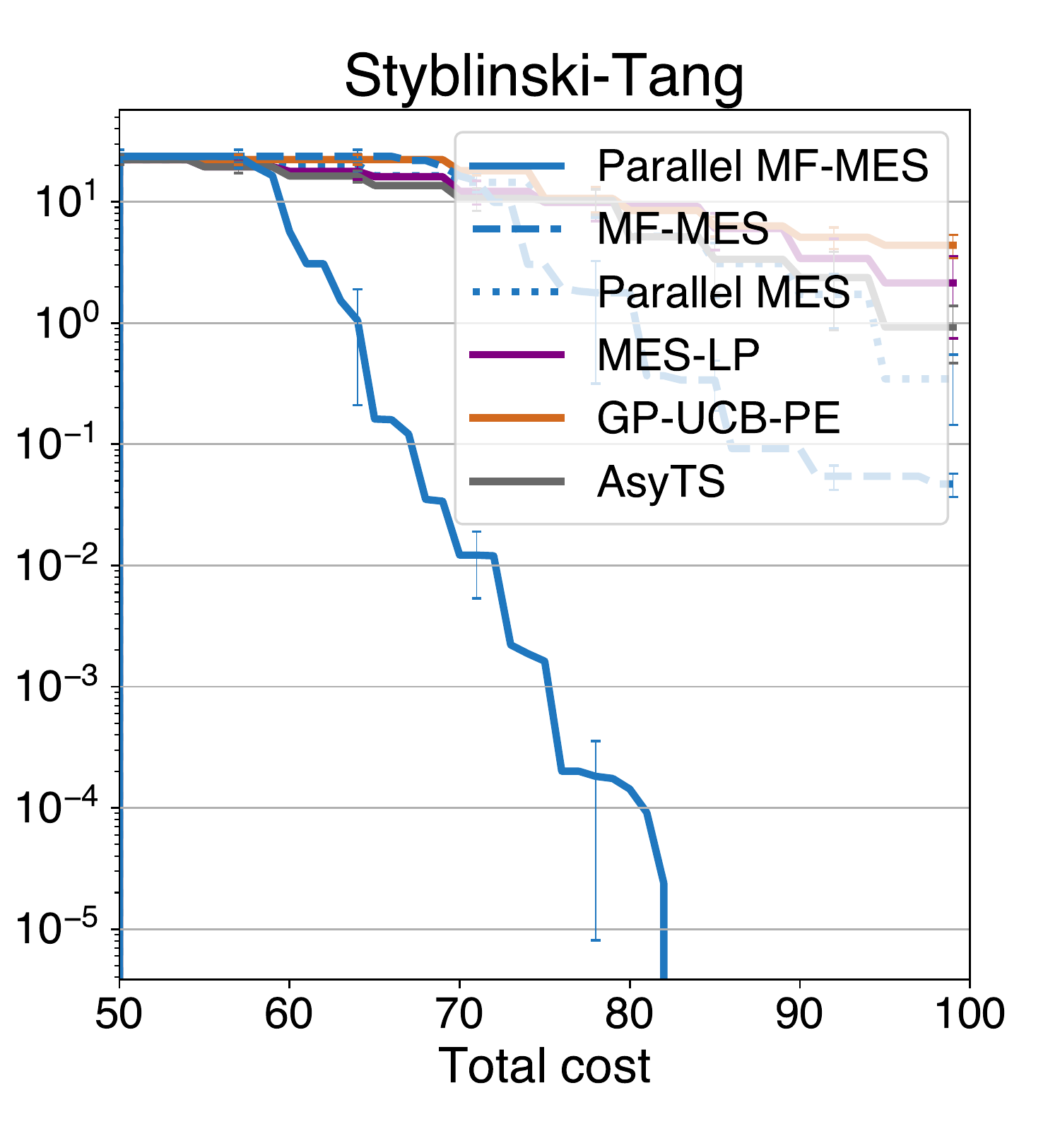}
\igr{.24}{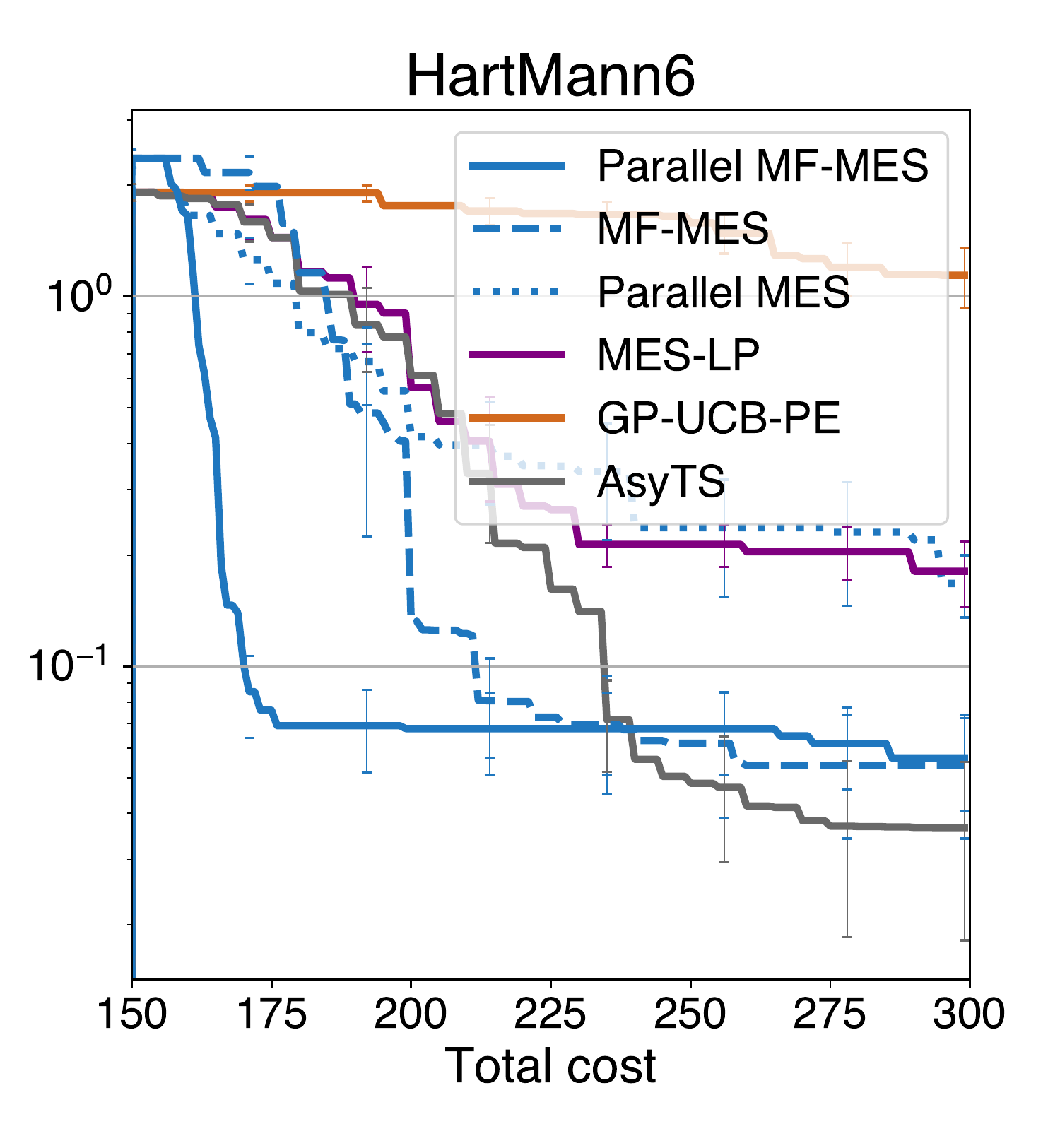}
\igr{.24}{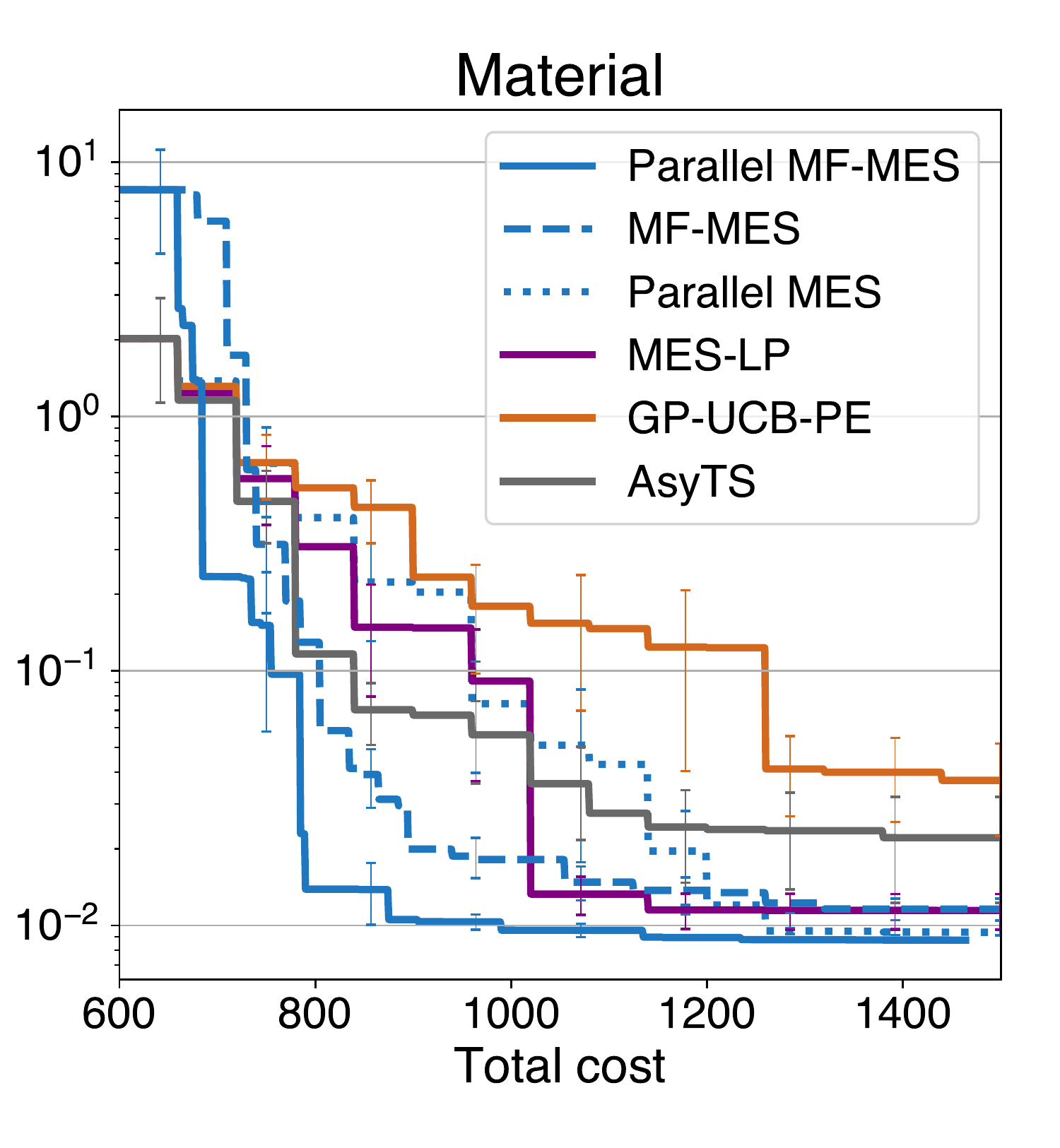}
}

\subfloat[Inference regret.]{
\igr{.24}{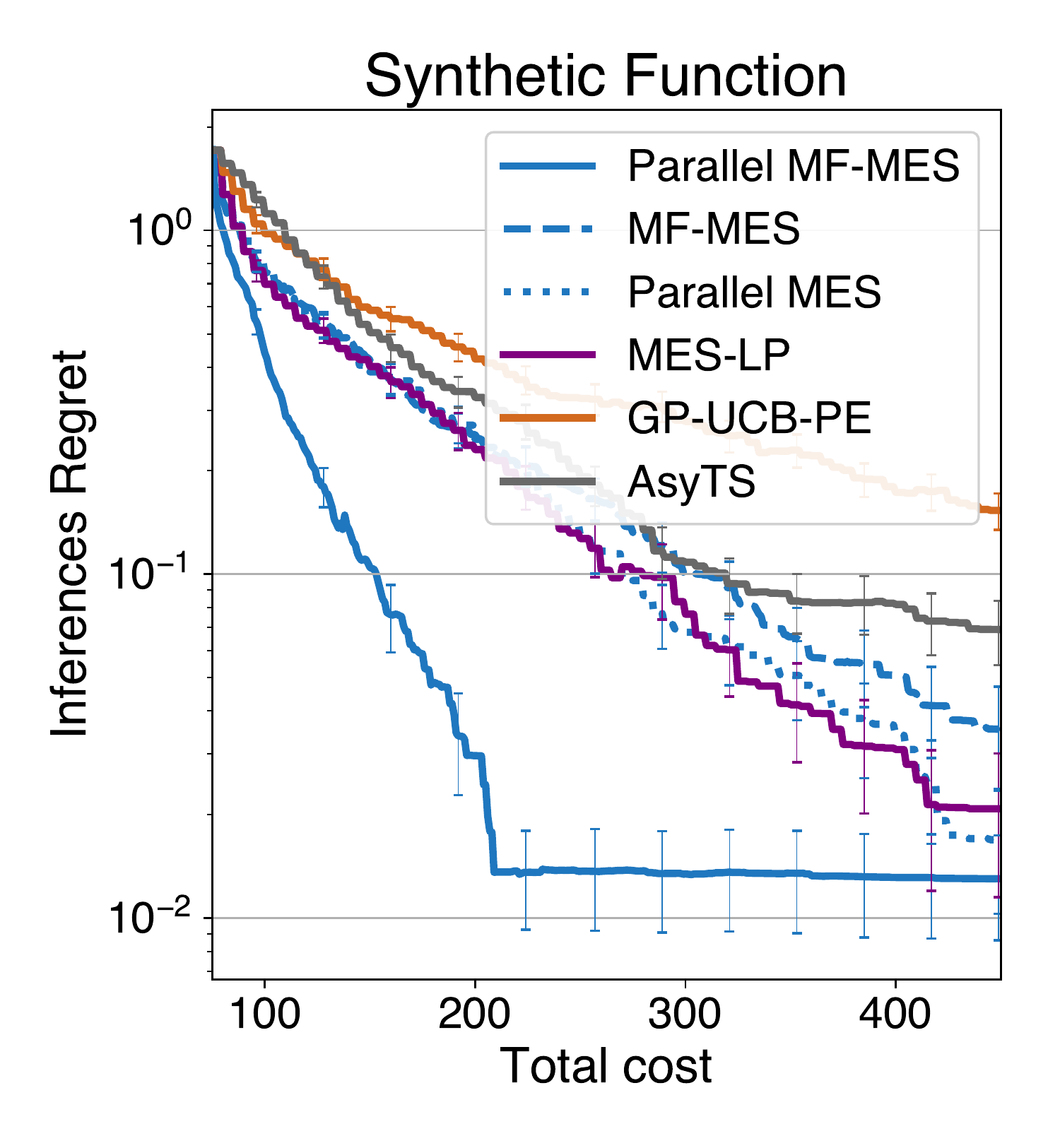}
\igr{.24}{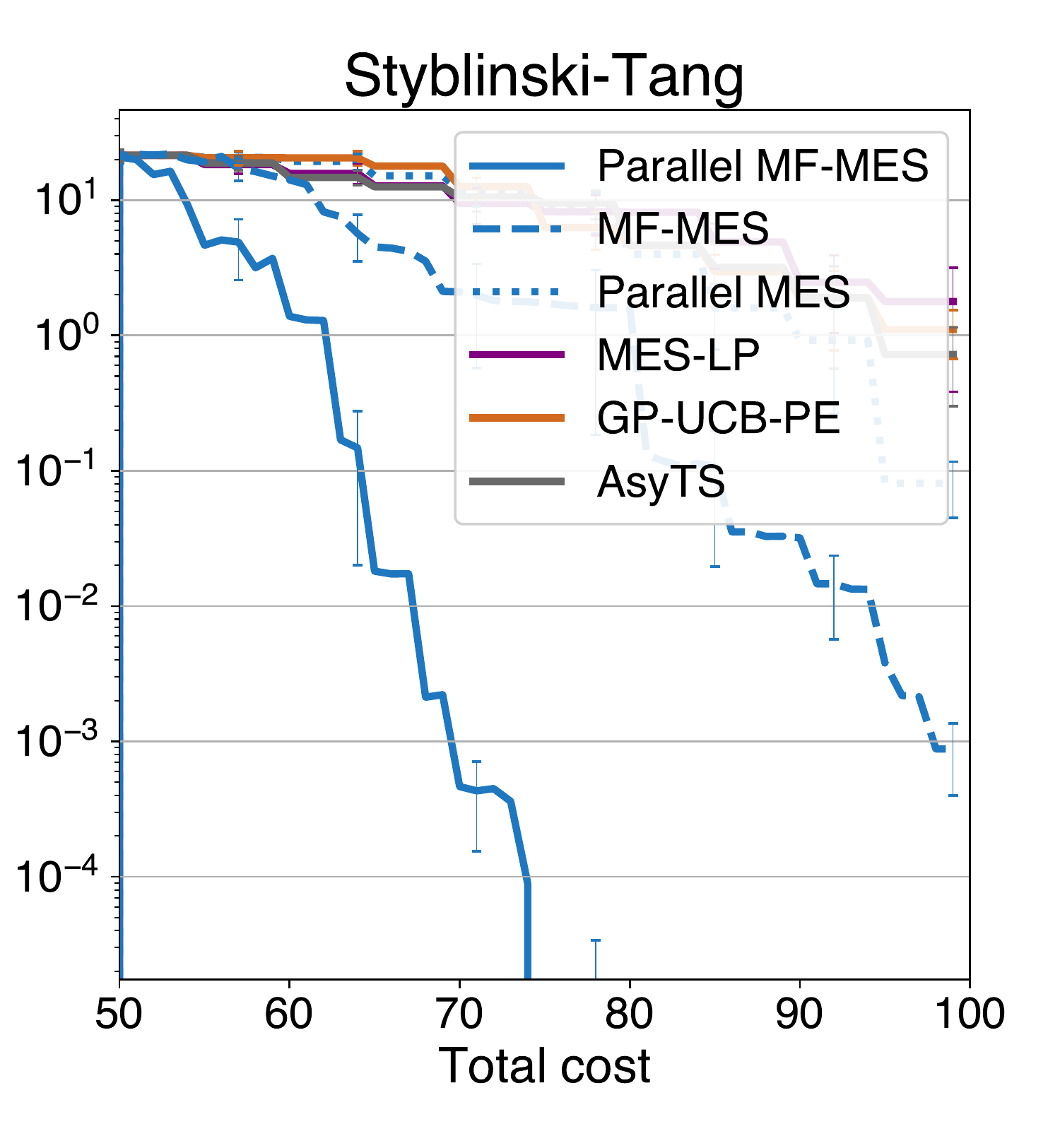}
\igr{.24}{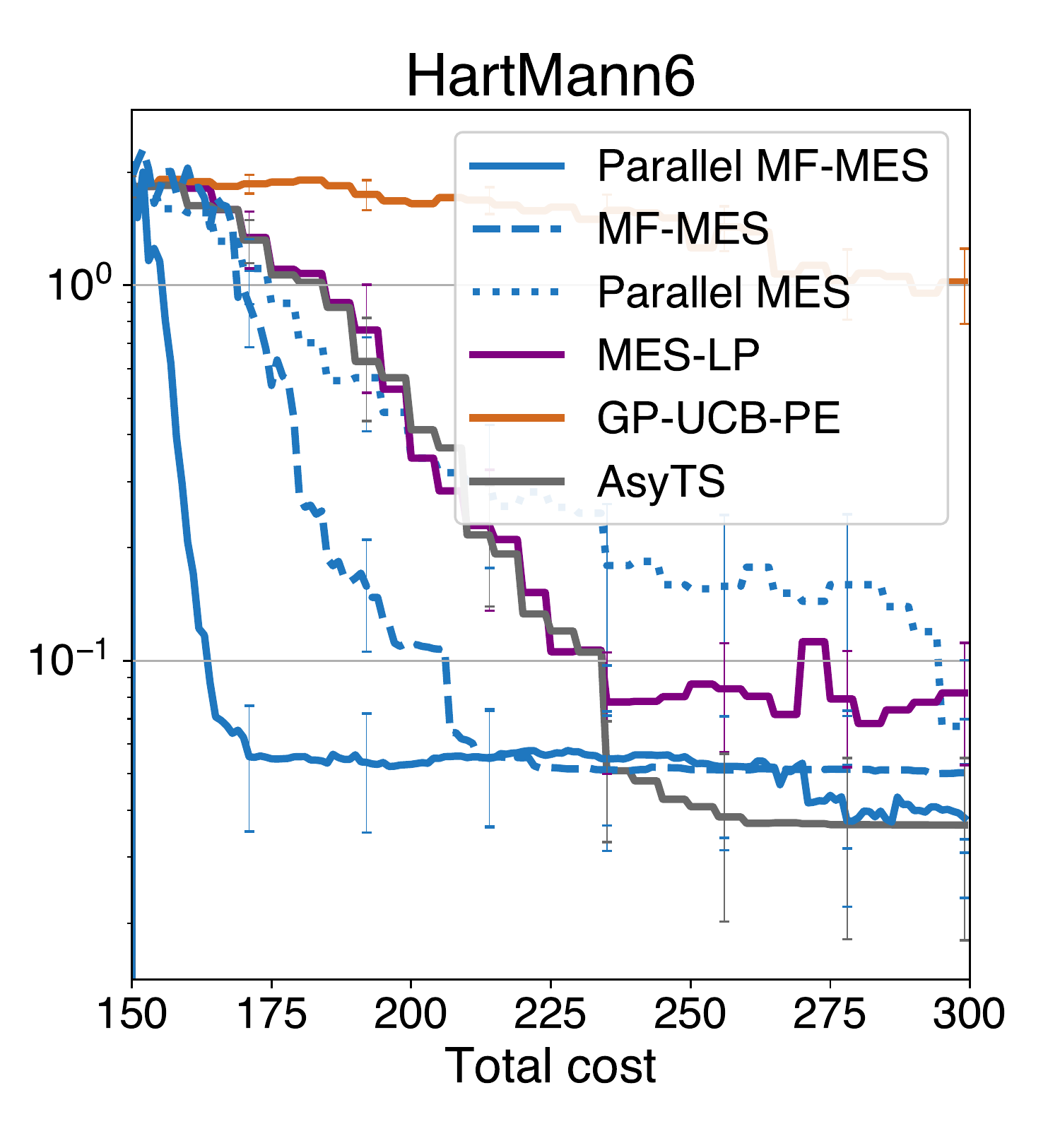}
\igr{.24}{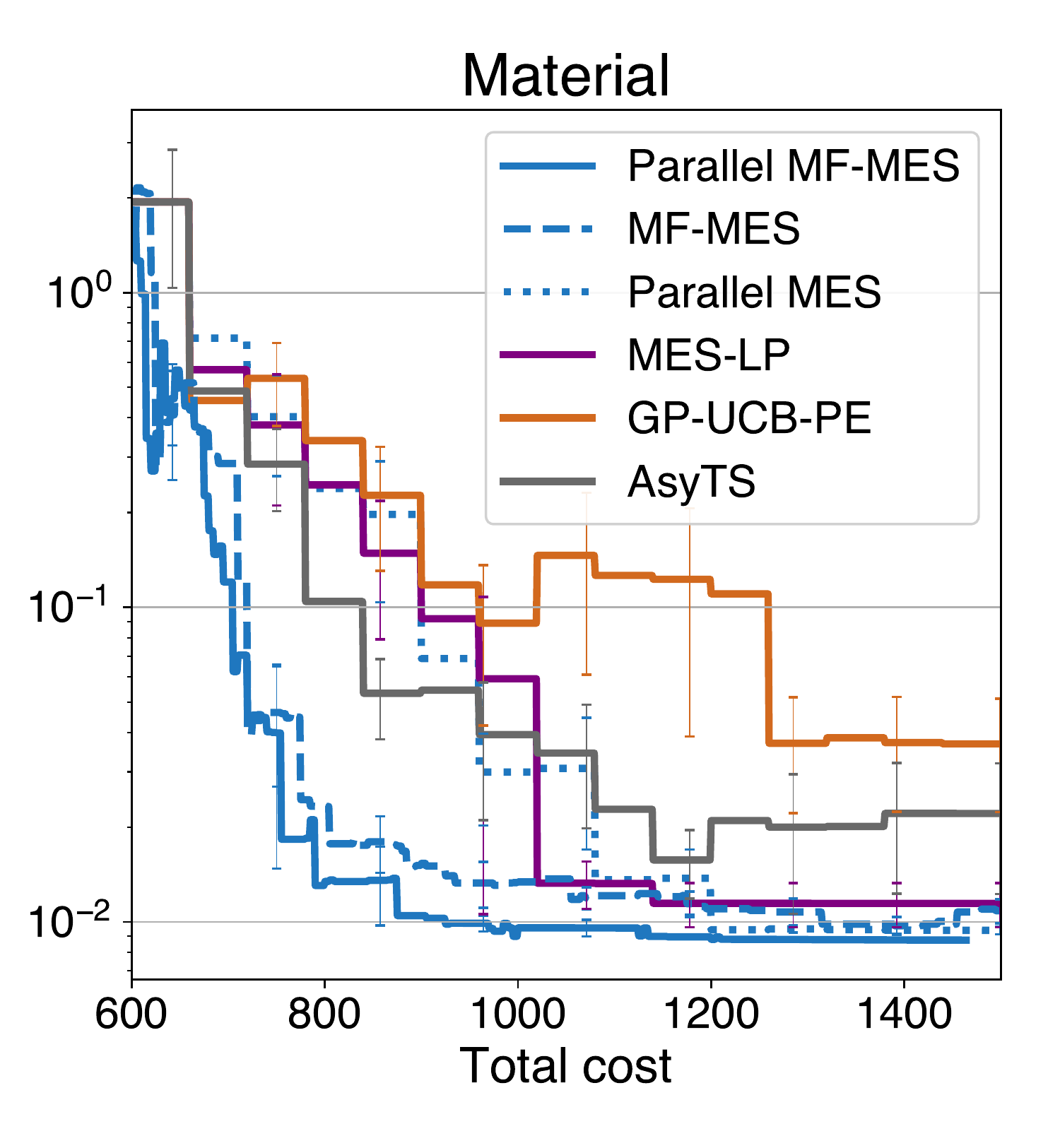}
}
\caption{Performance comparison on parallel querying.}
\label{fig:regret-parallel}
\end{figure*}

\subsection{Evaluation for Sequential Querying}
\label{sec:}

We first evaluate the performance for sequential querying.
For comparison, we used MF-SKO \citep{Huang2005-Sequential},
Bayesian optimization with continuous approximations (BOCA) \citep{Kandasamy2017-Multi},
and multi-fidelity PES (MF-PES) \citep{Zhang2017-Information}.
We also evaluated single fidelity MES which applied to the highest fidelity function $f^{(M)}(\*x)$.
As we see in Section~\ref{sec:related-work}, misoKG is another measure of global utility for MFBO.
However, we could not employ it as a baseline because it was not straightforward to modify the author implementation for fair comparison (e.g., changing the MF-GPR model), and creating efficient implementation from scratch is also extremely complicated (na{\"i}ve implementation of KG can be prohibitively slow).
Only BOCA employed the multi-task GPR (MT-GPR) model because the acquisition function assumes MT-GPR.
For the sampling of $f_*$ in MES and MF-MES, we employed the RFM-based approach described in Section~\ref{sec:alg}, and sampled $10$ $f_*$s at every iteration.
In MF-PES, $\*x_*$ was also sampled $10$ times through RFM as suggested by \citep{Hernandez2014-Predictive}.

\figurename~\ref{fig:regret-sequential} shows SR and IR.
In both of SR and IR, MF-MES decreased the regret faster than or comparable with all the other methods.
%
%
The single-fidelity MES is relatively slow because it cannot use lower-fidelity functions, and we clearly see that MF-MES successfully accelerates MES.
%
%
For SR of the GP-based synthetic, HartMann6 and material functions, MF-PES was slower than the others.
We empirically observed that MF-PES sometime did not aggressively select the highest fidelity samples enough.

A possible reason is in an approximation employed by MF-PES which assumes $f_{\*x}^{(m)} \leq f_{\*x_*}^{(m)} + c$ for $m < M$, where $c$ is a constant \citep[see][for the detailed definition]{Zhang2017-Information}.
However, even when $\*x_*$ is given, this strict inequality relation does not hold obviously (note that $\*x_*$ is the maximizer only when $m = M$), and we conjecture that the information gain from lower fidelity functions can be overly estimated because of this artificial truncation.
%
%
%
In the material data, IR was slightly unstable which was caused by noisy observations contained in this real-world dataset.
In particular, MF-PES largely fluctuated, and this would also be due to the lack of the highest fidelity samples as we mentioned above.
We also evaluate computational time of the acquisition functions in Appendix~\ref{app:time}.

\subsection{Evaluation for Parallel Querying}
\label{sec:}

Next, we evaluate performance on parallel querying.
For comparison, we used MES combined with local penalization \citep{Gonzalez2016-Batch}, denoted as MES-LP, Gaussian process upper confidence bound with pure exploration (GP-UCB-PE) \citep{Gonzalez2016-Batch}, asynchronous parallel Thompson sampling (AsyTS) \citep{Kandasamy2018-Parallelised}.
Here, we would like to note that no existing methods have been proposed for discrete fidelity parallel MFBO, to our knowledge, and extending existing methods to this setting is not straightforward because of discreteness of fidelity levels.
We also compare the performance of ``sequential'' MF-MES (which is same as ``MF-MES'' in \figurename~\ref{fig:regret-sequential}), and a parallel extension of single-fidelity MES (shown in Appendix~\ref{app:parallel-sync}) as baselines.
For the sampling of $\tilde{f}_*$ in Parallel MF-MES and Parallel MES, the number of samples are set $10$ through RFM.
The number of workers is set $q = 4$.

\figurename~\ref{fig:regret-parallel} shows SR and IR.
We see that parallel MF-MES substantially faster than sequential MF-MES and parallel MES.
This indicates that parallel MF-MES succeeded in assigning workers across multiple fidelities.
Compared with other methods, parallel MF-MES shows rapid or comparable convergence.

\section{Conclusion}
\label{sec:conclusion}

We propose a novel information-based multi-fidelity Bayesian optimization (MFBO).
The acquisition function is defined through the information gain for the optimal value $f_*$ of the highest fidelity function.
We show that our method called MF-MES (multi-fidelity max-value entropy search) can be reduced to simple computations, which allows reliable evaluation of the entropy.
For the asynchronous setting, which naturally arises in MFBO, we further propose parallelization of MF-MES and show that it is also easy to compute.
We demonstrate effectiveness of MF-MES by using benchmark functions and a real-world materials science data.

\clearpage

\section*{Acknowledgements}

This work was supported by MEXT KAKENHI to I.T. (16H06538, 17H00758), M.K. (16H06538, 17H04694) and M.S (16H02866); from JST CREST awarded to I.T. (JPMJCR1302, JPMJCR1502) and PRESTO awarded to M.K. (JPMJPR15N2), M.S (JPMJPR16N6) and Y.T (JPMJPR15NB); from the MI2I project of the Support Program for Starting Up Innovation Hub from JST awarded to I.T., and M.K.; and from RIKEN Center for AIP awarded to M.S. and I.T.

\clearpage





\bibliography{ref}
\bibliographystyle{stylename}

\clearpage



\appendix

\section{Semiparametric Latent Factor Model and its RFM approximation}
\label{app:SLFM}

\subsection{Model Definition}
\label{app:SLFM-model}

Semiparametric Latent Factor Model (SLFM) is a Gaussian process based multiple response model \citep{Teh2005-Semiparametric}.
SLFM represents each output as a sum of $C$ functions having different kernel functions $k_1, \ldots, k_C$, where
$k_c: \*x \times \*x \rightarrow \RR$
is a kernel function.
Let $w_{mc} \in \RR$ be a weight that the $m$-th output (fidelity) assigns to the $c$-th function.
By introducing an independent term
$\kappa_{cm} > 0$,
the kernel function is written as
\begin{align*}
 k((\*x,m),(\*x',m'))
 =
 \sum_{c = 1}^C (w_{cm} w_{cm'} + \kappa_{cm} \delta_{m = m'}) k_c(\*x,\*x'),
\end{align*}
where $\delta_{m = m'} = 1$ if $m = m'$, and $0$ otherwise.
The parameters
$w_{cm}$
and
$\kappa_{cm}$
which control dependence between multiple outputs are regarded as hyper-parameters, and standard approaches such as marginal likelihood optimization are often used to set them.

\subsection{RFM for SLFM}
\label{app:SLFM-RFM}

Let
$\*f_{\*x} \coloneqq (f^{(1)}_{\*x}, \ldots, f^{(M)}_{\*x})^\top$
be the $M$-dimensional output vector, and
\begin{align*}
 {\rm cov}(\*f_{\*x}, \*f_{\*x'})
 \coloneqq
 \begin{bmatrix}
  k((\*x,1),(\*x',1)) & \cdots & k((\*x,1),(\*x',M)) \\
  \vdots &   & \vdots   \\
  k((\*x,M),(\*x',1)) & \cdots & k((\*x,M),(\*x',M))
 \end{bmatrix}
\end{align*}
be the $M \times M$ covariance matrix of $\*x$ and $\*x'$.
By defining
$\*w_c \coloneqq (w_{c1}, \ldots, w_{cM})$
and
$\*\kappa_c \coloneqq (\kappa_{c1}, \ldots, \kappa_{cM})$,
this covariance is written as
\begin{align*}
 {\rm cov}(\*f_{\*x}, \*f_{\*x'})
 =
 \sum_{c = 1}^C
 (\*w_c \*w_c^\top + {\rm diag}(\*\kappa_c)) k_c(\*x,\*x').
\end{align*}
Since
$k_c(\*x,\*x')$
is assumed to be one of stationary kernel functions (e.g., Gaussian kernel), RFM can produce a feature vector representation $\*\phi_c$ which approximates the kernel function as
$k_c(\*x,\*x') \approx \*\phi^\top_c(\*x) \*\phi_c(\*x)$.
To transform
$\*w_c \*w_c^\top + {\rm diag}(\*\kappa_c)$
into a form of inner product, we use the Cholesky decomposition
\begin{align*}
 \*w_c \*w_c^\top + {\rm diag}(\*\kappa_c)
 =
 \*L_c
 \*L_c^\top,
\end{align*}
where $\*L_c \in \RR^{M \times M}$ is a lower triangular matrix.
Then, we obtain
\begin{align*}
 {\rm cov}(\*f_{\*x}, \*f_{\*x'})
 & \approx
 \sum_{c = 1}^C
 \*L_c  \*L_c^\top
 \left(
 \*\phi^\top_c(\*x) \*\phi_c(\*x')
 \right)
 \\ 
 & =
 \sum_{c = 1}^C
 \*\Psi^\top_c(\*x)
 \*\Psi_c(\*x')
\end{align*}
where
$\*\Psi_c(\*x) \coloneqq \*L_c^\top \otimes \*\phi_c(\*x)$.
Here, in the last line, we use the mixed-product property of Kronecker product.
Then, the $m$-th column of $\*\Psi_c(\*x)$ is defined as the feature of $\*x$ for the $m$-th fidelity
$\*\phi(\*x,m)$.

\section{Proof of Lemma~\ref{lemma:MFMES_conditional_entropy}}
\label{app:proof-lemma1}

%
Using Bayes' theorem, we obtain
\begin{align}
 &
 p(f^{(m)}_{\*x} \mid f^{(M)}_{\*x} \leq f_*, \cD_t)
 \nonumber \\ &
 =
 \frac
 {
 p(f^{(M)}_{\*x} \leq f_* \mid f^{(m)}_{\*x}, \cD_t ) p(f^{(m)}_{\*x} \mid \cD_t )
 }
 {
 p(f^{(M)}_{\*x} \leq f_* \mid \cD_t )
 }.
 \label{eq:bayes_theo}
\end{align}
The densities
$p(f^{(m)}_{\*x} \mid \cD_t )$
and
$p(f^{(M)}_{\*x} \leq f_* \mid \cD_t )$
are directly obtained from the predictive distribution:
\begin{align}
 \begin{split}
  p(f^{(m)}_{\*x} \mid \cD_t )
  &  =
  \phi(\gamma_{f^{(m)}_{\*x}}^{(m)} (\*x) ) / \sigma^{(m)}_{\*x},
  \\
  p(f^{(M)}_{\*x} \leq f_* \mid \cD_t )
  & =
  \Phi(\gamma_{f_*}^{(M)} (\*x)).
 \end{split}
 \label{eq:bayes_theo_sub}
\end{align}
In addition,  from \eq{eq:two-dim-conditional},
$p(f^{(M)}_{\*x} \leq f_* \mid f^{(m)}_{\*x}, \*x, \cD_t )$
is written as the cumulative distribution of this Gaussian:
\begin{align}
 p(f^{(M)}_{\*x} \leq f_* \mid f^{(m)}_{\*x}, \cD_t )
 =
 \Phi( (f_* - u(\*x)) / s(\*x) ). 
 \label{eq:cum_p_highf_given_lowf}
\end{align}
Substituting \eq{eq:bayes_theo_sub} and \eq{eq:cum_p_highf_given_lowf} into \eq{eq:bayes_theo}, the entropy is obtained.

\section{Information Gain with Noisy Observation}
\label{app:info-gain-with-noise}

Here, we describe calculation of the mutual information between $f_*$ and noisy observation $y^{(m)}_{\*x}$, where $y^{(m)}_{\*x} \coloneqq y^{(m)}(\*x)$ in this section.
The mutual information can be written as the difference of the entropy:
\begin{align}
 I(f_*; y^{(m)}_{\*x} \mid \*x, \cD_t)
 = H(y^{(m)}_{\*x} \mid \*x, \cD_t) - \EE_{p(f_* \mid\*x, \cD_t)}\bigl[ H(y^{(m)}_{\*x} \mid \*x, f_*, \cD_t)\bigl].
 \label{eq:mi_entropy_noise}
\end{align}
The first term in the right hand side is
\begin{align}
 H(y^{(m)}_{\*x} \mid \*x, \cD_t) =
 \log\left(
  \sqrt{ 2 \pi e (\sigma^{2(m)}_{\*x} + \sigma^2_{\mathrm{noise}}) }
 \right).
 \label{eq:H_y_noise}
\end{align}
Using the sampling approximation of $f_*$, the second term in (\ref{eq:mi_entropy_noise}) is
\begin{align}
  \EE_{p(f_* \mid\*x, \cD_t)}\bigl[ H(y^{(m)}_{\*x} \mid \*x, f_*, \cD_t)\bigl] \approx
 \sum_{f_* \in \cF_*}
 \frac{1}{ | \cF_* | }
 H(y^{(m)}_{\*x} \mid \*x, f_*, \cD_t).
 \label{eq:EH_y_noise}
\end{align}

For any $\zeta \in \RR$, define
\begin{align*}
\gamma^{(m)}_{\zeta}(\*x) &\coloneqq (\zeta - \mu^{(m)}_{\*x})/\sigma^{(m)}_{\*x},
\end{align*}
and
\begin{align*}
\rho^{(m)}_{\zeta}(\*x) &\coloneqq (\zeta - \mu^{(m)}_{\*x})/\sqrt{\sigma^{2(m)}_{\*x} + \sigma^2_{\mathrm{noise}}}.
\end{align*}
%
In this case, even for the highest fidelity $M$, the density
$p(y^{(m)}_{\*x} \mid \*x, f^{(M)}_{\*x} \leq f_*, \cD_t)$
is not the truncated normal because of the noise term.
%
Using Bayes' theorem, we decompose this density as
\begin{align}
 p(y^{(m)}_{\*x} \mid \*x, f^{(M)}_{\*x} \leq f_*, \cD_t)
  = \frac
 {
 p(f^{(M)}_{\*x} \leq f_* \mid y^{(m)}_{\*x}, \*x, \cD_t )
 p(y^{(m)}_{\*x} \mid \*x, \cD_t )
 }
 { p(f^{(M)}_{\*x} \leq f_* \mid \*x, \cD_t ) }.
 \label{eq:bayes_theo_noise}
\end{align}
%
%
The densities
$p(y^{(m)}_{\*x} \mid \*x, \cD_t )$
and
$p(f^{(M)}_{\*x} \leq f_* \mid \*x, \cD_t )$
are directly obtained from the predictive distribution:
\begin{align}
 \begin{split}
  & p(y^{(m)}_{\*x} \mid \*x, \cD_t )
  = \frac{1}{\sqrt{\sigma^{2(m)}_{\*x} + \sigma^2_{\mathrm{noise}}}} \phi(\rho_{y^{(m)}_{\*x}}^{(m)} (\*x)),\\
  &
  p(f^{(M)}_{\*x} \leq f_* \mid \*x, \cD_t ) =
  \Phi(\gamma_{f_*}^{(M)}(\*x)).
 \end{split}
 \label{eq:bayes_theo_sub_noise}
\end{align}
The joint marginal distribution
$p(f^{(M)}_{\*x}, y^{(m)}_{\*x}  \mid \*x, \cD_t )$
is written as
\begin{align*}
 \begin{split}
  &
  \begin{bmatrix}
   y^{(m)}_{\*x} \\
   f^{(M)}_{\*x}
  \end{bmatrix}
  \mid \*x, \cD_t
   \sim
  \cN
  \left(
  \begin{bmatrix}
   \mu^{(m)}_{\*x} \\
   \mu^{(M)}_{\*x}
  \end{bmatrix},
  \begin{bmatrix}
   {\sigma^2}^{(m)}_{\*x} + \sigma^2_{\mathrm{noise}} & {\sigma^2}^{(mM)}_{\*x} \\
   {\sigma^2}^{(mM)}_{\*x}           & {\sigma^2}^{(M)}_{\*x}
  \end{bmatrix}
  \right),
 \end{split}
\end{align*}
From this distribution, we obtain
$p(f^{(M)}_{\*x} \mid y^{(m)}_{\*x}, \*x, \cD_t )$
as
\begin{align*}
 f^{(M)}_{\*x} \mid y^{(m)}_{\*x}, \*x, \mathcal{D}_t
 \sim
 \mathcal{N}(u_{\rm noise}(\*x), s^2_{\rm noise}(\*x)),
\end{align*}
where
\begin{align*}
 u_{\rm noise}(\*x)
 & =
 \frac{{\sigma^2}^{(mM)}_{\*x} \bigl(y^{(m)}_{\*x}- \mu^{(m)}_{\*x}\bigl)}{{\sigma^2}^{(m)}_{\*x} + \sigma^2_{\mathrm{noise}}}  + \mu^{(M)}_{\*x},
 \\
 s^2_{\rm noise}(\*x)
 & =
 {\sigma^2}^{(M)}_{\*x} - \frac{\bigl({\sigma^2}^{(mM)}_{\*x} \bigl)^2}{{\sigma^2}^{(m)}_{\*x} + \sigma^2_{\mathrm{noise}}}.
\end{align*}
Thus,
$p(f^{(M)}_{\*x} \leq f_* \mid y^{(m)}_{\*x}, \*x, \cD_t )$
is written as the cumulative distribution of this Gaussian:
\begin{align}
 p(f^{(M)}_{\*x} \leq f_* \mid y^{(m)}_{\*x}, \*x, \cD_t )
 =
 \Phi( \gamma^{\prime}_{f_*}(\*x) ),
 \label{eq:cum_p_highf_given_lowf_noise}
\end{align}
where,
$\gamma^{\prime}_{f_*}(\*x) \coloneqq (f_* - u_{\rm noise}(\*x)) / s_{\rm noise}(\*x)$.
%
Using \eq{eq:bayes_theo}, \eq{eq:bayes_theo_sub}, and \eq{eq:cum_p_highf_given_lowf} in the proof of Lemma~\ref{lemma:MFMES_conditional_entropy}, the entropy is obtained as
\begin{align}
   H(y^{(m)}_{\*x} & \mid  \*x, f^{(M)}_{\*x} \leq f_*, \cD_t) \nonumber \\
    &= - \int Z {\Phi\bigl(\gamma^{\prime}_{f_*}(\*x)\bigl) \phi\bigl(\rho^{(m)}_{y^{(m)}_{\*x}}(\*x)\bigl)}
     \cdot \log
   \left(
   Z
   {\Phi\bigl(\gamma^{\prime}_{f_*}(\*x)\bigl) \phi\bigl(\rho^{(m)}_{y^{(m)}_{\*x}}(\*x)\bigl)}
   \right)
   \mathrm{d} y^{(m)}_{\*x},
  \label{eq:H_lowf_trunc_highf_noise}
  \end{align}
where
 $Z \coloneqq
 { 1 } /
 { \sqrt{\sigma^{2(m)}_{\*x} + \sigma^2_{\mathrm{noise}}}  \Phi(\gamma^{(M)}_{f_*}(\*x)) }$.
%
The integral in \eq{eq:H_lowf_trunc_highf_noise} can be calculated by using numerical integration in the same way as \eq{eq:H_lowf_trunc_highf}.
%


Using
$I(f_* ; y^{(m)}_{\*x})$
instead of
$I(f_* ; f^{(m)}_{\*x})$
would be more natural when the observations are assumed to contain the observation noise with large variance $\sigma^2_{\rm noise}$, but in practice, difference of these two formulations would not largely effect on performance of BO when $\sigma^2_{\rm noise}$ is small.

Note that the mutual information of parallel querying
$I(f_* ; f_{\*x}^{(m)} \mid \cD_t, \*f_{\cQ})$
can be replaced with the noisy observation
$I(f_* ; y_{\*x}^{(m)} \mid \cD_t, \*f_{\cQ})$
by using same procedure.

\section{Additional Information for Parallel Querying}

\subsection{Proof of Lemma~\ref{lemma:paraMFMES_minfo}}
\label{app:proof-para}


The first term of \eq{eq:minfo_para} is
\begin{align*}
 \EE_{\*f_{\cQ} \mid \cD_t}
 \left[
 H(f_{\*x}^{(m)} \mid \cD_t, \*f_{\cQ})
 \right]
 & =
 \EE_{\*f_{\cQ} \mid \cD_t}
 \left[
 \log
 \left(
 \sigma_{\*x \mid \*f_{\cQ}}^{(m)} \sqrt{2 \pi e}
 \right)
 \right]
 \\
 & =
 \log
 \left(
 \sigma_{\*x \mid \*f_{\cQ}}^{(m)} \sqrt{2 \pi e}
 \right).
\end{align*}
The last equation holds since
$\sigma_{\*x \mid \*f_{\cQ}}^{(m)}$
does not depend on
$\*f_{\cQ}$.

The second term of \eq{eq:minfo_para} is written as
\begin{align}
 &
 \EE_{\*f_{\cQ}, f_* \mid \cD_t }
 \left[
 H(f_{\*x}^{(m)} \mid \cD_t, \*f_{\cQ}, f_{\*x}^{(M)} \leq f_*)
 \right]
 \nonumber
 \\
 & =
 -
 \int
 \int
 p(\*f_{\cQ}, f_* \mid \cD_t)
 \int
 p(f_{\*x}^{(m)} \mid \cD_t, \*f_{\cQ}, f_{\*x}^{(M)} \leq f_*)
 \log
 p(f_{\*x}^{(m)} \mid \cD_t, \*f_{\cQ}, f_{\*x}^{(M)} \leq f_*)
 {\rm d} f_{\*x}^{(m)}
 {\rm d} \*f_{\cQ}
 {\rm d} f_*.
 \label{eq:minfo_para_second}
\end{align}


For the conditional distribution
\begin{align*}
 f_{\*x}^{(M)} \mid \cD_t, \*f_{\cQ}, f_{\*x}^{(m)}
 \sim
 \cN(u_p(\*x), s_p^2(\*x)),
\end{align*}
the mean and the variance function can be written as
\begin{align*}
 u_p(\*x)
 & =
 \frac{
 \sigma^{2 (m M)}_{\*x \mid \*f_{\cQ}}
 \left(
 f_{\*x}^{(m)} -  \mu_{\*x \mid \*f_{\cQ}}^{(m)}
 \right)
 }{
 \sigma^{2 (m)}_{\*x \mid \*f_{\cQ}}
 }
 +
 \mu^{(M)}_{\*x \mid \*f_{\cQ}},
 \\ 
 s_p^2(\*x)
 & =
 \sigma^{2 (M)}_{\*x \mid \*f_{\cQ}}
 -
 \left(
 \sigma_{\*x \mid \*f_{\cQ}}^{2 (m M)}
 \right)^2
 ~ / ~
 \sigma_{\*x \mid \*f_{\cQ}}^{2 (m)}.
\end{align*}
Then, from Bayes' theorem, we see
\begin{align}
 p(f_{\*x}^{(m)} \mid \cD_t, \*f_{\cQ}, f_{\*x}^{(M)} \leq f_*)
 & =
 \frac
 {
 p(f_{\*x}^{(M)} \leq f_* \mid \cD_t, f_{\*x}^{(m)}, \*f_{\cQ})  p(f_{\*x}^{(m)} \mid \cD_t, \*f_{\cQ})
 }
 {
 p(f_{\*x}^{(M)} \leq f_* \mid \cD_t, \*f_{\cQ})
 }
 \nonumber
 \\ 
 & =
 \frac
 {
 \Phi \left( \frac{ f_* - u_p(\*x) }{ s_p(\*x) } \right)
 \phi \left( \frac{ f^{(m)}_{\*x} - \mu^{(m)}_{\*x \mid \*f_{\cQ}} }{ \sigma^{(m)}_{\*x \mid \*f_{\cQ}} } \right)
 }
 {
 \sigma^{(m)}_{\*x \mid \*f_{\cQ}}
 \Phi \left( \frac{ f_* - \mu^{(M)}_{\*x \mid \*f_{\cQ}} }{ \sigma^{(M)}_{\*x \mid \*f_{\cQ}} } \right)
 }.
 \label{eq:bayes_theo_para}
\end{align}
By defining
\begin{align*}
 A
 \coloneqq
 \frac{
 \sigma^{2 (m M)}_{\*x \mid \*f_{\cQ}}
 }{
 \sigma^{2 (m)}_{\*x \mid \*f_{\cQ}}
 },
\end{align*}
we can re-write
\begin{align*}
 \tilde{f}_* - u_p(\*x) = \tilde{f}_* - A \tilde{f}_{\*x}^{(m)},
\end{align*}
and then, \eq{eq:bayes_theo_para} is transformed into
\begin{align*}
 \frac{
 \Phi\left(
 \frac{
 \tilde{f}_* - A \tilde{f}^{(m)}_{\*x}
 }
 {
 s_p(\*x)
 }
 \right)
 \phi\left(\frac{ \tilde{f}^{(m)}_{\*x} }{ \sigma^{(m)}_{\*x \mid \*f_{\cQ}}} \right)
 }
 {
 \sigma^{(m)}_{\*x \mid \*f_{\cQ}}
 \Phi\left( \frac{ \tilde{f}_* }{ \sigma^{(M)}_{\*x \mid \*f_{\cQ}} } \right)
 }
 \eqqcolon
 \eta(\tilde{f}_*, \tilde{f}_{\*x}^{(m)}).
\end{align*}
By further defining
$h(\tilde{f}_*, \tilde{f}_{\*x}^{(m)}) \coloneqq \eta(\tilde{f}_*, \tilde{f}_{\*x}^{(m)}) \log \eta(\tilde{f}_*, \tilde{f}_{\*x}^{(m)})$,
we simplify
\eq{eq:minfo_para_second}
as follows
\begin{align}
 -
 \int
 \int
 p(\*f_{\cQ}, f_* \mid \cD_t)
 \int
 h(\tilde{f}_*, \tilde{f}_{\*x}^{(m)})
 {\rm d} f_{\*x}^{(m)}
 {\rm d} \*f_{\cQ}
 {\rm d} f_*.
 \label{eq:minfo_para_second_2}
\end{align}
This indicates that the most inner integrand can be shown as a function which only depends two random variables
$\tilde{f}_*$ and $\tilde{f}_{\*x}^{(m)}$.
We change the variables of integration from
$(f_*, f_{\*x}^{(m)}, \*f_{\cQ}^\top)^\top$
to
$(\tilde{f}_*, \tilde{f}_{\*x}^{(m)}, \*f_{\cQ}^\top)^\top$.

\begin{align*}
 \cJ
 &
 \coloneqq
 \begin{bmatrix}
  \pd{ \tilde{f}_* }{ f_* }
  &
  \pd{ \tilde{f}_* }{ f_{\*x}^{(m)} }
  &
  \pd{ \tilde{f}_* }{ \*f^\top_{\cQ} }
  \\ 
  \pd{ \tilde{f}_{\*x}^{(m)} }{ f_* }
  &
  \pd{ \tilde{f}_{\*x}^{(m)} }{ f_{\*x}^{(m)}  }
  &
  \pd{ \tilde{f}_{\*x}^{(m)} }{ \*f^\top_{\cQ} }
  \\ 
  \pd{ \*f_{\cQ} }{ f_* }
  &
  \pd{ \*f_{\cQ} }{ f_{\*x}^{(m)} }
  &
  \pd{ \*f_{\cQ} }{ \*f^\top_{\cQ} }
 \end{bmatrix}
 \\
 &
 =
 \begin{bmatrix}
  \*I_{2}
  &
  \*\Sigma_{\cM,\cQ}
  \*\Sigma_{\cQ}^{-1}
  \\ 
  \*0
  &
  \*I_{|\cQ|}
 \end{bmatrix}
\end{align*}
where
$\*I_{2}$
and
$\*I_{|\cQ|}$
are the identity matrices with size $2$ and $|\cQ|$, respectively.
Note that determinant of $\cJ$ is $| \cJ | = 1$.
Thus, by changing variables of integration and variables of the densities, \eq{eq:minfo_para_second_2} can be transformed into
\begin{align}
 -
 \int
 \int
 p(\*f_{\cQ}, f_* \mid \cD_t)
 \int
 h(\tilde{f}_*, \tilde{f}_{\*x}^{(m)})
 {\rm d} f_{\*x}^{(m)}
 {\rm d} \*f_{\cQ}
 {\rm d} f_*
 &
 =
 -
 \int
 \int
 p(\*f_{\cQ}, \tilde{f}_* \mid \cD_t)
 \int
 h(\tilde{f}_*, \tilde{f}_{\*x}^{(m)})
 {\rm d} \tilde{f}_{\*x}^{(m)}
 {\rm d} \*f_{\cQ}
 {\rm d} \tilde{f}_*
 \nonumber
 \\
 &
 =
 -
 \int
 p(\tilde{f}_* \mid \cD_t)
 \int
 h(\tilde{f}_*, \tilde{f}_{\*x}^{(m)})
 {\rm d} \tilde{f}_{\*x}^{(m)}
 {\rm d} \tilde{f}_*
 \nonumber
 \\
 &
 =
 -
 \EE_{\tilde{f}_* \mid \cD_t}
 \left[
 \int
 h(\tilde{f}_*, \tilde{f}_{\*x}^{(m)})
 {\rm d} \tilde{f}_{\*x}^{(m)}
 \right].
 \label{eq:minfo_para_second_3}
\end{align}

\subsection{Analytical Calculation of Entropy for $m = M$}
\label{app:analytic-entropy-para}

When $m = M$, the most inner integral in \eq{eq:minfo_para_second} can be further simplified because it is equal to the entropy of the truncated normal
$p(f_{\*x}^{(M)} \mid \cD_t, \*f_{\cQ}, f_{\*x}^{(M)} \leq f_*)$, which is written as
\begin{align*}
&
-\int p(f_{\*x}^{(M)} \mid \cD_t, \*f_{\cQ}, f_{\*x}^{(M)} \leq f_*) \log p(f_{\*x}^{(M)} \mid \cD_t, \*f_{\cQ}, f_{\*x}^{(M)} \leq f_*) {\rm d} f_{\*x}^{(M)}
\\
& =
\log \left(
 \sqrt{2 \pi e } \sigma^{(M)}_{\*x \mid \*f_{\cQ}}
 \Phi \left(
 \frac{ \tilde{f}_* }{ \sigma^{(M)}_{\*x \mid \*f_{\cQ}}  }
 \right)
\right)
-
\frac{ \tilde{f}_* }{ \sigma^{(M)}_{\*x \mid \*f_{\cQ}}  }
\frac
{
 \phi \left(
 \frac{ \tilde{f}_* }{ \sigma^{(M)}_{\*x \mid \*f_{\cQ}}  }
 \right)
}
{
 2 \Phi \left(
 \frac{ \tilde{f}_* }{ \sigma^{(M)}_{\*x \mid \*f_{\cQ}}  }
 \right)
}
\\
&
\eqqcolon
\omega(\tilde{f}_*)
,
\end{align*}
%
%
By using the same change of variables as \eq{eq:minfo_para_second_3}, we obtain
\begin{align*}
 -
 \int
 \int
 p(\*f_{\cQ}, f_* \mid \cD_t)
 \omega(\tilde{f}_*)
 {\rm d} \*f_{\cQ}
 {\rm d} f_*
 =
 -
 \EE_{\tilde{f}_* \mid \cD_t}
 \left[
 \omega(\tilde{f}_*)
 \right].
\end{align*}


\subsection{Algorithm}
\label{app:alg-para}

As shown in Algorithm~\ref{alg:MF-MES-para}, the acquisition function maximization is performed when a worker becomes available.
The sampling of $\tilde{f}_* \in \widetilde{\cF}_*$ is performed through an RFM approximation of MF-GPR: $\*w^\top \*\phi(\*x,m)$.
For the entropy calculation in line 19, one dimensional numerical integration is necessary for the integral in \eq{eq:H1_para} when $m \neq M$, while the analytical formula is available when $m = M$ as shown in Appendix~\ref{app:analytic-entropy-para}.

\begin{algorithm}[t]
 \caption{Parallel MF-MES}
 \label{alg:MF-MES-para}
 \begin{algorithmic}[1]
  \Function{Parallel MF-MES}{$\cD_0, M, \cX, \{ \lambda^{(m)} \}_{m=1}^M$}
  \For{$t = 0, \ldots, T$}
  \State Wait for a worker to be available
  \State Generate $\widetilde{\cF}_*$ from RFM
  \State $(\*x_{t+1}, m_{t+1}) \leftarrow \argmax_{\*x \in \cX, m}$
  \par  \hskip2em
  {\textproc{InfoGain}}($\*x$, $m$, $\widetilde{\cF}_*$, $\cD_t$) $/$ $\lambda^{(m)}$
  \State $\cD_{t+1} \leftarrow \cD_{t} \cup (\*x_{t+1}, y^{(m_{t+1})}(\*x_{t+1}), m_{t+1})$
  \EndFor
  \EndFunction
  \Function{InfoGain}{$\*x$, $m$, $\cF_*$, $\cD_t$}
  \State Calculate $\mu^{(m)}_{\*x \mid \*f_{\cQ}}$ and $\sigma^{(m)}_{\*x \mid \*f_{\cQ}}$
  \State Set $H_0 \leftarrow \log \left( \sigma_{\*x \mid \*f_{\cQ}} \sqrt{2 \pi e} \right)$
  \If{$m \neq M$}
  \State Calculate $\mu^{(M)}_{\*x \mid \*f_{\cQ}}, \sigma^{(m)}_{\*x \mid \*f_{\cQ}}$, and $\sigma^{2(mM)}_{\*x \mid \*f_{\cQ}}$
  \EndIf
  \State Set $H_1 \leftarrow$ \eq{eq:H1_para}
  \State Return $H_0 - H_1$
  \EndFunction
 \end{algorithmic}
\end{algorithm}

\subsection{Synchronous Parallelization}
\label{app:parallel-sync}

\subsubsection{Single-fidelity Setting}
\label{app:parallel-sync-sf}

\begin{figure}[t]
 \centering
 \igr{0.6}{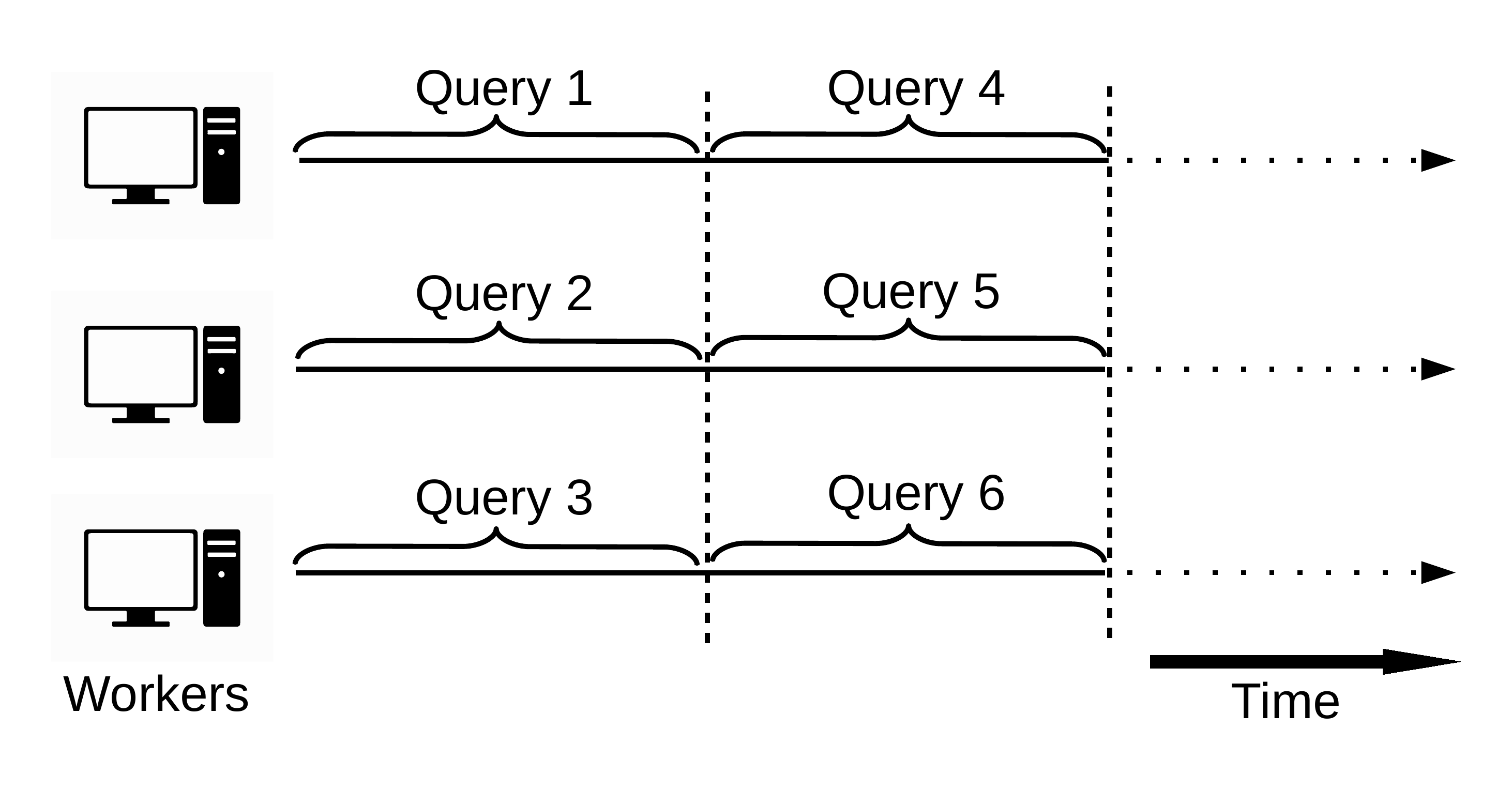}
 \caption{Synchronous setting in parallel BO.}
 \label{fig:syncBO}
\end{figure}

In the main text, we focus on the asynchronous setting because of the diversity of sampling costs in MFBO.
On the other hand, many parallel BO studies on the single-fidelity setting consider the synchronous setting (\figurename~\ref{fig:syncBO}).
To our knowledge, a parallel extension of MES has not been studies even in the single-fidelity setting.
Our approach is actually applicable to defining the single fidelity acquisition function.
Although our main focus is in MFBO, we here show a counterpart of our multi-fidelity acquisition function in the single fidelity setting.

Suppose that we need to select $q$ points written as
$\cQ = \{\*x_1, \ldots, \*x_q \}$
for the single fidelity parallel BO.
Unlike the asynchronous setting, $q$ points is needed to be selected simultaneously.
By setting
$\*f_{\cQ} \coloneqq (f_{\*x_1}, \ldots, f_{\*x_q})^\top$,
a natural extension of MES for synchronous single-fidelity setting is written as
\begin{align}
 I(f_* ; \*f_{\cQ} \mid \cD_t)
 \coloneqq
 H(\*f_{\cQ} \mid \cD_t)
 -
 \EE_{\*f_{\cQ} \mid \cD_t}
 \left[
 H(\*f_{\cQ} \mid \*f_{\cQ} \leq f_*, \cD_t)
 \right].
 \label{eq:minfo_sync}
\end{align}
Note that we impose the condition $\*f_{\cQ} \leq f_*$, indicating that all the elements of $\*f_{\cQ}$ is less than or equal to $f_*$, instead of $f_{\*x} \leq f_*$ in the usual MES.
The first term is the entropy of the $q$-dimensional Gaussian distribution which can be analytically calculated.
The second term is the entropy of the multi-variate truncated normal distribution, for which we show analytical and approximate approaches to the computation.

First, we consider the analytical approach.
The density
$p(\*f_{\cQ} \mid \cD_t)$
is the predictive distribution of GPR, and we define $\*\mu_{\cQ}$ and $\*\Sigma_{\cQ}$ as the mean and covariance matrix, respectively.
The truncated normal in the second term is defined through this density as follows
\begin{align}
 p(\*f_{\cQ} \mid \*f_{\cQ} \leq f_*, \cD_t)
 =
 \begin{cases}
  p(\*f_{\cQ} \mid \cD_t) / Z, & \text{ if } \*f_{\cQ} \leq f_*, \\
  0, & \text{ otherwise, }
 \end{cases}
 \label{eq:TNQ}
\end{align}
where
\begin{align*}
 Z \coloneqq
 \int_{\*f_{\cQ} \leq f_*}
 p(\*f_{\cQ} \mid \cD_t)
 \mathrm{d} \*f_{\cQ}.
\end{align*}
We refer to the truncated normal \eq{eq:TNQ} as
${\rm TN}(\*\mu^{\rm TN}_{\cQ}, \*\Sigma^{\rm TN}_{\cQ})$,
where $\*\mu^{\rm TN}_{\cQ}$ and $\*\Sigma^{\rm TN}_{\cQ}$ are the mean and covariance matrix, respectively.
Let
$\EE_{\rm TN}$
be the expectation by the density \eq{eq:TNQ}.
Then, the entropy in the second term of \eq{eq:minfo_sync} is re-written as
\begin{align*}
 H[\bm{f}_{\cQ} \mid \mathcal{D}, \bm{f}_{\cQ} \leq f_*]
 &= -
 \int_{\*f_{\cQ} \leq f_*}
 \frac{p(\*f_{\cQ} \mid \cD_t)}{Z} \log \frac{p(\*f_{\cQ} \mid \cD_t)}{Z} \mathrm{d}\bm{f}_{\cQ} \\
 &= - \mathbb{E}_{\mathrm{TN}}\biggl[ \log \frac{p(\*f_{\cQ} \mid \cD_t)}{Z} \biggl] \\
 &= - \mathbb{E}_{\mathrm{TN}}\bigl[ \log p(\*f_{\cQ} \mid \cD_t) - \log Z \bigl] \\
 &= - \mathbb{E}_{\mathrm{TN}}\bigl[ \log p(\*f_{\cQ} \mid \cD_t) \bigl] + \log Z \\
 &= - \mathbb{E}_{\mathrm{TN}}\biggl[ -\frac{1}{2} \log |2 \pi \*\Sigma_{\cQ}| - \frac{1}{2} (\bm{f}_{\cQ} - \bm{\mu}_{\cQ})^\top \*\Sigma_{\cQ}^{-1} (\bm{f}_{\cQ} - \bm{\mu}_{\cQ}) \biggl] + \log Z \\
 &= \frac{1}{2} \log |2 \pi \*\Sigma_{\cQ}| + \frac{1}{2} \underbrace{\mathbb{E}_{\mathrm{TN}}\biggl[(\bm{f}_{\cQ} - \bm{\mu}_{\cQ})^\top \*\Sigma_{\cQ}^{-1} (\bm{f}_{\cQ} - \bm{\mu}_{\cQ}) \biggl]}_{\eqqcolon B} + \log Z.
\end{align*}
By defining $\bm{d} = \bm{\mu}_{\cQ}^{\mathrm{TN}} - \bm{\mu}_{\cQ}$, we see
\begin{align*}
 B
 &= \mathbb{E}_{\mathrm{TN}}\bigl[ \mathrm{Tr}\bigl(\*\Sigma_{\cQ}^{-1} (\bm{f}_{\cQ} - \bm{\mu}_{\cQ}) (\bm{f}_{\cQ} - \bm{\mu}_{\cQ})^\top \bigl) \bigl] \\
 &= \mathrm{Tr}\bigl( \*\Sigma_{\cQ}^{-1} \mathbb{E}_{\mathrm{TN}}\bigl[(\bm{f}_{\cQ} - \bm{\mu}_{\cQ}) (\bm{f}_{\cQ} - \bm{\mu}_{\cQ})^\top \bigl] \bigl) \\
 &= \mathrm{Tr}\bigl( \*\Sigma_{\cQ}^{-1} \mathbb{E}_{\mathrm{TN}}\bigl[(\bm{f}_{\cQ} - \*\mu_{\cQ}^{\mathrm{TN}} + \bm{d}) (\bm{f}_{\cQ} - \*\mu_{\cQ}^{\mathrm{TN}} + \bm{d})^\top \bigl] \bigl) \\
 &= \mathrm{Tr}\bigl( \*\Sigma_{\cQ}^{-1} \mathbb{E}_{\mathrm{TN}}\bigl[(\bm{f}_{\cQ} - \*\mu_{\cQ}^{\mathrm{TN}}) (\bm{f}_{\cQ} - \*\mu_{\cQ}^{\mathrm{TN}})^\top
 + \bm{d}(\bm{f}_{\cQ} - \bm{\mu}_{\cQ}^{\rm TN})^\top + (\bm{f}_{\cQ} - \bm{\mu}_{\cQ}^{\rm TN}) \bm{d}^\top + \bm{d}\bm{d}^\top \bigl] \bigl). \\
\end{align*}
Since $\mathbb{E}_{\mathrm{TN}}[(\bm{f}_{\cQ} - \bm{\mu}_{\cQ})] = \bm{0}$, we further obtain
\begin{align*}
    B
    &= \mathrm{Tr}\bigl( \*\Sigma_{\cQ}^{-1} \mathbb{E}_{\mathrm{TN}}\bigl[(\bm{f}_{\cQ} - \*\mu_{\cQ}^{\mathrm{TN}}) (\bm{f}_{\cQ} - \*\mu_{\cQ}^{\mathrm{TN}})^\top + \bm{d}\bm{d}^\top \bigl] \bigl) \\
    &= \mathrm{Tr}\bigl( \*\Sigma_{\cQ}^{-1} ( \*\Sigma_{\cQ}^{\mathrm{TN}} + \bm{d}\bm{d}^\top ) \bigl) \\
\end{align*}
Therefore, 
we obtain
\begin{align*}
 H[\bm{f}_Q \mid \mathcal{D}, \bm{f}_Q \leq f_*]
 &= \frac{1}{2} \Bigl( \log |2 \pi \bm{\Sigma}_{\cQ}| + \mathrm{Tr}\bigl( \bm{\Sigma}_{\cQ}^{-1} ( \bm{\Sigma}_{\cQ}^{\mathrm{TN}} + \bm{d}\bm{d}^\top ) \bigl) \Bigl)  + \log Z.
\end{align*}
If $Z$, $\*\mu_{\cQ}^{\rm TN}$, and $\*\Sigma_{\cQ}^{\rm TN}$ are available, the above equation is easily calculated.
%
%
The normalization term $Z$ is the $q$-dimensional Gaussian CDF, for which a lot of fast computation algorithms have been proposed \citep[e.g.,][]{Genz1992-numerical,Genton2017-Hierarchical}.
A method proposed by \citep{Genz1992-numerical} has been widely used, which requires $O(q^2)$ computations.
%
%
For
$\*\mu_{\cQ}^{\rm TN}$, and $\*\Sigma_{\cQ}^{\rm TN}$,
\citet{Manjunath2012-moments} shows analytical formulas which also depend on the multivariate Gaussian CDF.
This needs $q$ times computations of the $q-1$ dimensional CDF, and $q(q-1)$ times computations of the $q-2$ dimensional CDF.

To avoid many computations of $q-1$ dimensional CDF, we can introduce approximation of the entropy calculation or greedy selection of $\cQ$.
As a fast approximation, expectation propagation (EP) can be used to replace the truncated normal distribution with a Gaussian distribution, which makes the entropy calculation analytical.
The similar technique is also used in \citep{Hernandez2014-Predictive}.
For the greedy strategy, we can choose a next point to add $\cQ$ by maximizing
$I(f_* ; \*f_{\*x} \mid \cD_t, \*f_{\tilde{\cQ}})$, where
$\tilde{\cQ}$
is a set of $(\*x,m)$ already determined to be included in $\cQ$.
This information can be evaluated by the same way as we saw in the asynchronous setting \eq{eq:minfo_para} because the equation has the same form of conditional mutual information.

\subsubsection{Multi-fidelity Setting}
\label{app:parallel-sync-mf}

Combining the synchronous setting with multi-fidelity functions $m = 1, \ldots, M$ results in a combinatorial selection of
$\cQ = \{ (\*x_1,m_1), \ldots, (\*x_q,m_q) \}$
because of the discreteness of the fidelity level $m$.
When a simple greedy strategy is employed to select $\cQ$, the procedure is reduced to the almost the same procedure as the synchronous single fidelity case described above.
This indicates that we can avoid the $q$ dimensional integral by using the technique shown in Section~\ref{sec:parallel}.

\section{Incorporating Fidelity Feature}
\label{app:extension-z-space}

%
Our proposed method is applicable to the case that the fidelity is defined as a point of a fidelity feature (FF) space $\cZ$ instead of the discrete fidelity level $1, \ldots, M$ \citep{Kandasamy2017-Multi}.
Let $f^{(\*z)}_{\*x}$ be the predictive distribution for the fidelity $\*z \in \cZ$.
The goal is to solve $\max_{\*x \in \cX} f^{(\*z_*)}_{\*x}$, where $\*z_* \in \cZ$ is the highest fidelity to be optimized.
For example, in the neural network hyper-parameter optimization, $\cZ$ can be a two dimensional space defined by the number of training data and the number of training iterations.

In this case, our acquisition function
\eq{eq:acq}
is extended to
\begin{align}
 a(\*x,\*z)
 \coloneqq
 {I(f_* ; f^{(\*z)}_{\*x})}
 ~ / ~
 {\lambda^{(\*z)}},
 \label{eq:mutual_info_z}
\end{align}
where $f_* \coloneqq \max_{\*x \in \cX} f^{(\*z_*)}_{\*x}$ in this case, and $\lambda^{(\*z)}$ is known cost for $\*z \in \cZ$.
As with \citep{Kandasamy2017-Multi}, we represent the output $f^{(\*z)}_{\*x}$ as a Gaussian process on the direct product space $\cX \times \cZ$.
%
%
Suppose that the observed training data set is written as
$\cD_{n} = \{ (\*x_i, y^{(\*z_i)}(\*x_i), \*z_i) \}_{i=1}^n$,
where $y^{(\*z_i)}(\*x_i)$ is an observation of $\*x_i$ at the fidelity $\*z_i$.
A standard approach to defining a kernel on the joint space $\cX \times \cZ$ is to use the product form
$k((\*x_i,\*z_i),(\*x_j,\*z_j)) = k_x(\*x_i,\*x_j) ~ k_z(\*z_i,\*z_j)$,
where
$k_x: \cX \times \cX \rightarrow \RR$
is a kernel for the input space $\cX$, and
$k_z: \cZ \times \cZ \rightarrow \RR$
is a kernel for the fidelity space $\cZ$.
Based on this kernel, predictive distribution of GPR can be defined for any pair of $(\*x, \*z)$, and thus the numerator of (\ref{eq:mutual_info_z}) can be calculated by using the same approach as $I(f_* ; f^{(m)}_{\*x})$ which we describe in Section~\ref{sec:info-gain}.

Parallelization can also be considered in this FF-based case.
For the asynchronous setting, the acquisition function is
\begin{align*}
 a_{\rm para}(\*x,\*z) =
 I(f_* ; f_{\*x}^{(\*z)} \mid \cD_t, \*f_{\cQ})
 / \lambda^{(\*z)},
\end{align*}
in which information gain is conditioned on the set of points currently under evaluation $\cQ = \{ (\*x_1, m_1), \dots, \linebreak (\*x_{q-1}, m_{q-1}) \}$.
As in the sequential case above, the calculation of this acquisition function is almost same as the discrete case in Section~\ref{sec:parallel}.
For the synchronous case, the same discussion as Appendix~\ref{app:parallel-sync} also holds.

\section{Summary of Settings in Sequential/Parallel MFBO}
\label{app:summary-of-settings}

A possible combination of the single/multiple fidelity and sequential/parallel querying are summarized in \tablename~\ref{tab:summary}.
Our main focus is in FF-free MFBO, and FF-free parallel MFBO with asynchronous querying.
In particular, for parallel MFBO, except for the FF-based synchronous querying, no prior works exist to our knowledge.

\begin{table}[t]
 \caption{
 Summary of possible settings.
 ``FF-based'' indicates the setting that the fidelity feature $z$ is available, while ``FF-free'' does not assume it.
 Synchronous querying is denoted as 'sync', and asynchronous querying is denoted as 'asyn'.
 }
 \label{tab:summary}
 \centering
 { \footnotesize
\begin{tabular}{r|l|l|l|l|}
& Fidelity & (S)equential/ & Our description & Note \\
&          & (P)arallel    &                 & \\ \hline
Parallel BO & Single & P (sync) & Appendix~\ref{app:parallel-sync-sf} &  - \\
            & Single & P (asyn) & Special case of Parallel MF-MES     &  - \\
\hdashline
MFBO        & Multiple (FF-based) & S & Appendix~\ref{app:extension-z-space}  & - \\
            & Multiple (FF-free) & S & MF-MES described in Section~\ref{sec:info-gain} & - \\
\hdashline
Parallel MFBO & Multiple (FF-based) & P (sync) & Appendix~\ref{app:extension-z-space} & \citep{Wu2017-Continuous} \\ 
              & Multiple (FF-based) & P (asyn) & Appendix~\ref{app:extension-z-space} & No prior work \\
              & Multiple (FF-free) & P (sync) & Appendix~\ref{app:parallel-sync-mf} & No prior work \\
              & Multiple (FF-free) & P (asyn) & Parallel MF-MES described in Section~\ref{sec:parallel} & No prior work
\end{tabular}
 }
\end{table}


\section{Additional Information of Empirical Evaluation}
\label{}

\subsection{Other Experimental Settings}
\label{app:settings}

\subsubsection{Settings of Methods}

We trained the GPR model using normalized training observations (mean $0$, and standard deviation $1$), other than the GP-based synthetic function.
Model hyper-parameters were optimized by marginal-likelihood at every 5 iterations.
For the GP-based synthetic function, we set the GPR hyper-parameters as parameters used for sampling the function.
For the initial observations, we employed the Latin hypercube approach shown by \citep{Huang2005-Sequential}.
The number of initial training points $\bm{x} \in \mathcal{X} \subset \mathbb{R}^d$ were set as follows:
\begin{itemize}
 \item $5d$ and $4d$ for $m =1$ and $2$, respectively, if $M = 2$
 \item $6d$, $3d$ and $2d$ for $m = 1, 2$ and $3$, respectively, if $M = 3$
 \item $10d$, $7d$ and $3d$ for $m = 1, 2$ and $3$, respectively, in the material dataset
\end{itemize}
We used the Gaussian kernel
$k(\bm{x}, \bm{x}^\prime) = \exp(- \sum_{i=1}^{d}( \*x_i - \*x_i^\prime )^2 / (2 \ell_i^2))$
for all kernels.
The length scale parameter $\ell_d$ was optimized through marginal-likelihood in the following interval:
\begin{itemize}
 \item $\ell_d \in [\mathrm{Domain\ size} / 10, \mathrm{Domain\ size} \times 10]$
       for the GP-based synthetic function and the benchmark functions, here $\mathrm{Domain\ size}$ is the difference between the maximum and the minimum of the input domain in each dimension.
       The input domain of each function is shown in Appendix~\ref{app:settings-benchmark}.
 \item $\ell_d \in [10^{-3}, 10^{-1}]$ for the material dateset
 \item The task kernel in BOCA: $\ell_d \in [2, (M-1)\times 10]$ for benchmark functions, and $\ell_d \in [10, 10^3]$ for the material dataset
\end{itemize}
The noise parameter of GPR was fixed as $\sigma^2_{\mathrm{noise}} = 10^{-6}$.
The number of kernels in SLFM was $C=2$.
The hyper-parameters in covariance among different output dimension were also optimized through marginal-likelihood in the following interval:
\begin{itemize}
   \item $w_{c1} \in [\sqrt{0.75}, 1]$ for $c= 1, 2$
   \item $w_{c2} \in [-\sqrt{0.25}, \sqrt{0.25}]$ for $c= 1, 2$
   \item $\kappa_{cm} \in [10^{-3}, 10^{-1}]$ for $c=1, 2$ and $m = 1, \dots, M$
\end{itemize}
The number of basis $D$ in RFM was $1000$, which was used by MF-MES, MF-PES, MES-LP, and AsyTS.
The number of samplings for $f^*$ in MES and PES was $10$.


For all compared methods, including BOCA, MFSKO, local penalization in MES-LP, GP-UCB-PE, and AsyTS, we followed the settings of hyper-parameters in their original papers.


\subsubsection{Details of Benchmark Datasets}
\label{app:settings-benchmark}

\paragraph{GP-based Synthetic functions}

We used RFM for SLFM described in Appendix~\ref{app:SLFM-RFM}.
The input dimension is $d = 3$ and the domain is $x_i \in [0, 1]$.
The parameters are
$C=1, \*w = (0.9, 0.9)^\top, \*\kappa = (0.1, 0.1)^\top$,
and
$\ell_i = 0.1$ for $i = 1, 2, 3$.




\paragraph{Styblinski-Tang function}

\begin{align*}
    f^{(1)} &= \frac{1}{2} \sum_{i=1}^2 (0.9 x_i^4-15x_i^2+6x_i), \\
    f^{(2)} &= \frac{1}{2} \sum_{i=1}^2 (x_i^4-16x_i^2+5x_i), \\
    x_i &\in [-5, 5], i=1, 2
\end{align*}

\paragraph{HartMann6 function}

\begin{align*}
  f^{(1)} &= - \sum_{i=1}^4 (\alpha_i - 0.2) \exp \biggl( -\sum_{j=1}^6 A_{ij} (x_j - P_{ij})^2 \biggl), \\
  f^{(2)} &= - \sum_{i=1}^4 (\alpha_i - 0.1) \exp \biggl( -\sum_{j=1}^6 A_{ij} (x_j - P_{ij})^2 \biggl)\\
  f^{(3)} &= - \sum_{i=1}^4 \alpha_i \exp \biggl( -\sum_{j=1}^6 A_{ij} (x_j - P_{ij})^2 \biggl)\\
  \bm{\alpha} &= [1.0, 1.2, 3.0, 3.2]^\top \\
  \bm{A} &= \left( \begin{array}{cccccc}
      10 & 3 & 17 & 3.5 & 1.7 & 8 \\
      0.05 & 10 & 17 & 0.1 & 8 & 14 \\
      3 & 3.5 & 1.7 & 10 & 17 & 8 \\
      17 & 8 & 0.05 & 10 & 0.1 & 14
  \end{array}\right) \\
  \bm{P} &= 10^{-4} \left( \begin{array}{cccccc}
      1312 & 1696 & 5569 & 124 & 8283 & 5886 \\
      2329 & 4135 & 8307 & 3736 & 1004 & 9991 \\
      2348 & 1451 & 3522 & 2883 & 3047 & 6650 \\
      4047 & 8828 & 8732 & 5743 & 1091 & 381
  \end{array}\right) \\
  \bm{x}_j &\in [0, 1], j = 1, \dots, 6
\end{align*}

\paragraph{Materials Data}

As an example of practical application, we applied our method to the parameter optimization of computational simulation model in materials science. 
There is a computational model \citep{Tsukada2014-Equilibrium} that predicts equilibrium shape of precipitates in the $\alpha$-Mg phase when material parameters are given.
We 
estimate two material parameters (lattice mismatch and interface energy between the $\alpha$-Mg and precipitate phases) from experimental data on precipitate shape measured by transmission electron microscopy (TEM) \citep{Bhattacharjee2013-effect}.
The objective function is the discrepancy between precipitate shape predicted by the computational model and one measured by TEM.
\subsection{Measuring Computational Time of Acquisition Functions}
\label{app:time}

We measured the computational time for the maximization of the acquisition functions.
We assume that the predictive distribution of the GPR model is already obtained, because it is almost common for all the methods.
The training dataset is created by the initialization process in our experiment described in Appendix~\ref{app:settings}.

\figurename~\ref{fig:time-comparison} shows the results on three benchmark dataset, used in the main text.
BOCA and MFSKO are relatively easy to compute because they are based on UCB and EI, respectively.
Their acquisition function is simple, but difficult to incorporate global utility of the candidate without tuning parameters as we discuss in the main text.
MF-MES was much faster than MF-PES.
We emphasize that MF-PES employs the approximation based on EP to accelerate the computation, unlike our MF-MES which is almost analytical.
This indicates that MF-MES provides more reliable entropy computation with smaller amount of computations than MF-PES.

\begin{figure}[t]
 \centering
 \subfloat[Styblinski-Tang]{
 \igr{0.4}{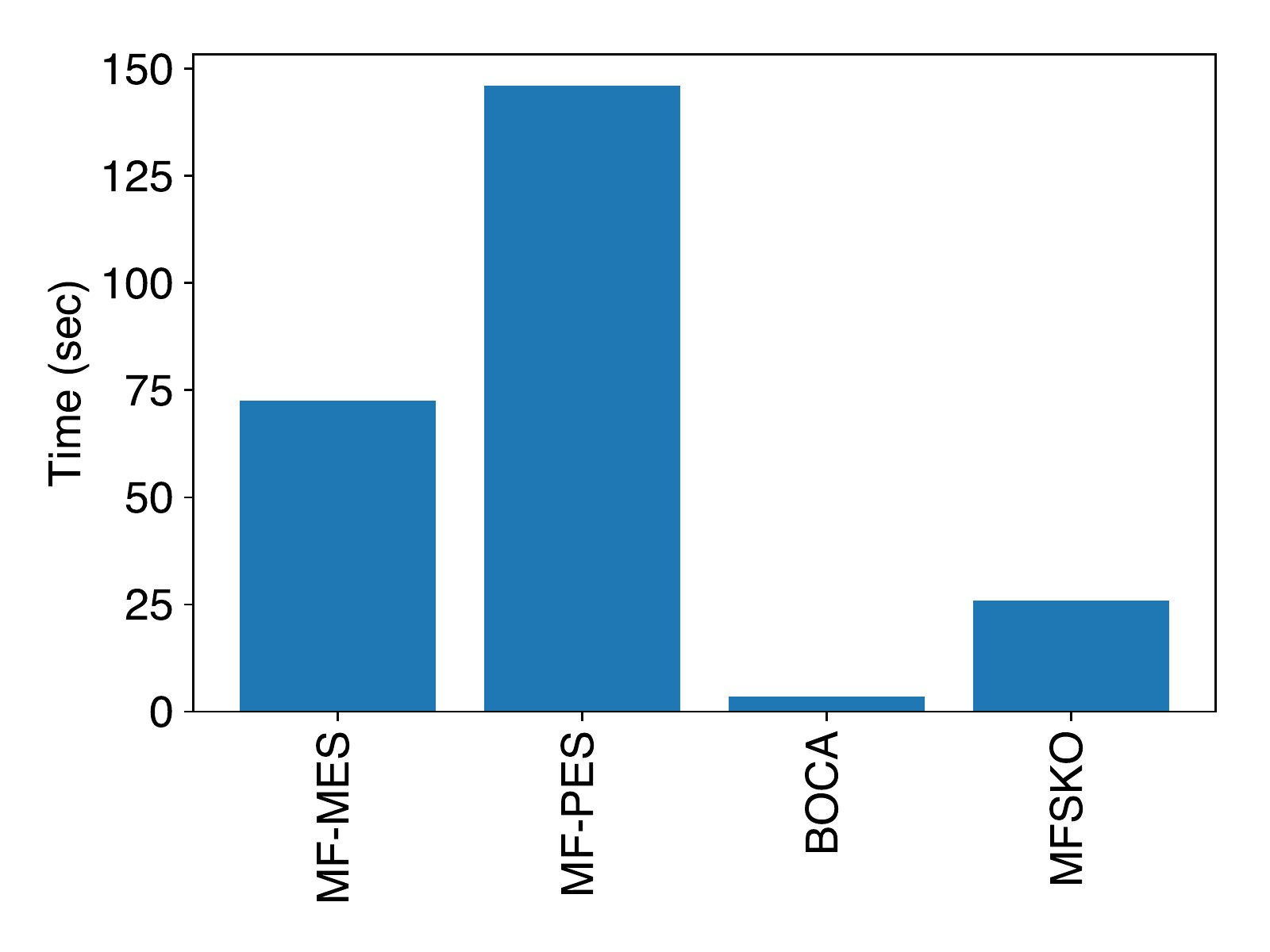}
 }
 \subfloat[HartMann6]{
 \igr{0.4}{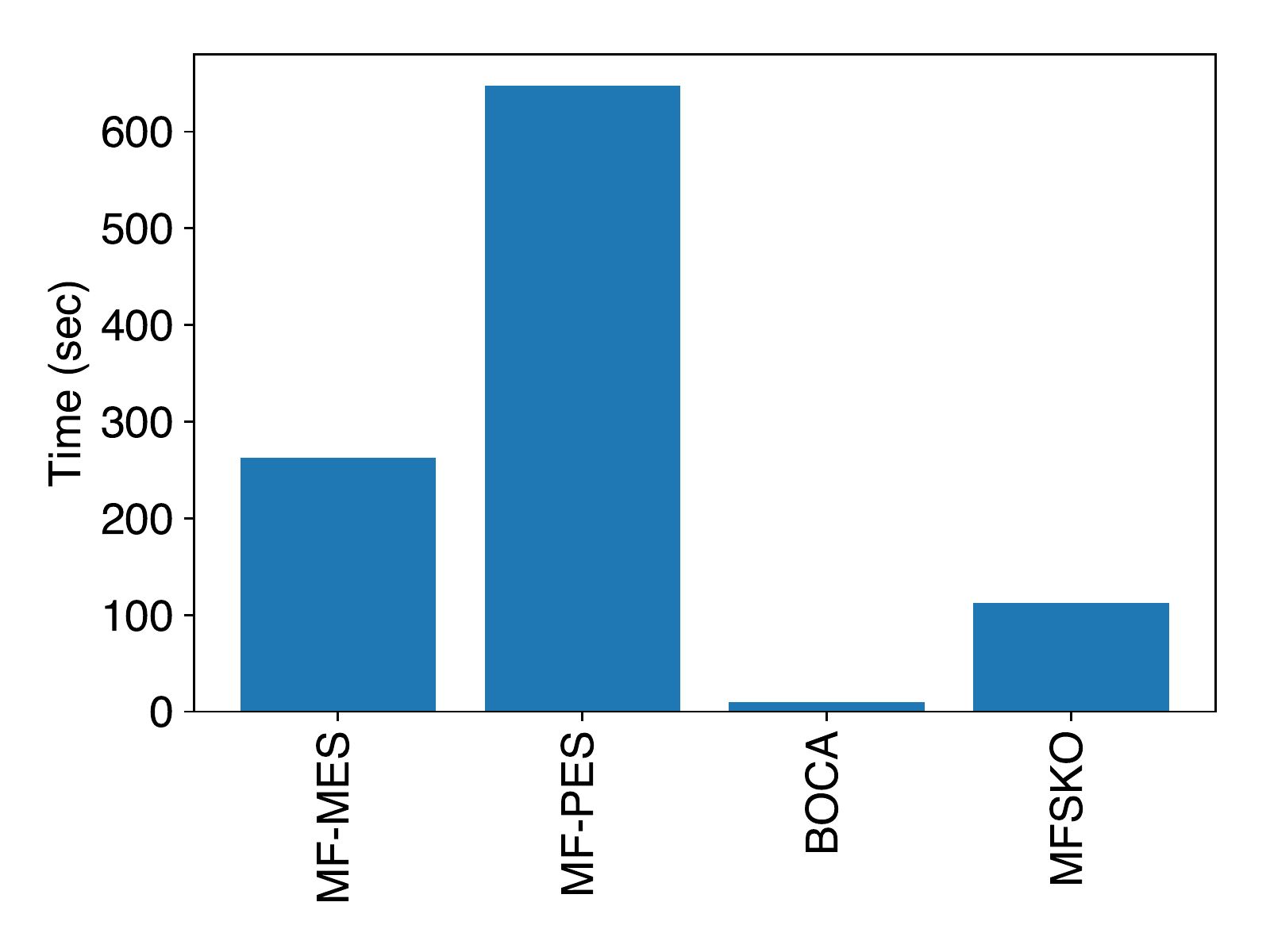}
 }
 \caption{
 Computational time for acquisition function maximization.
 }
 \label{fig:time-comparison}
\end{figure}

\end{document}

%% file: packages.tex
\usepackage{microtype}
\usepackage[dvipdfmx]{graphicx}
\usepackage{booktabs} 

\usepackage{amsmath,amssymb,ascmac,amsthm}

\usepackage{mediabb}
\usepackage{bm}

\usepackage{xcolor}
\usepackage{here}
\usepackage{natbib}
\usepackage{listings}
\usepackage{authblk}
\usepackage{mathtools}
\usepackage{subfig}
\usepackage{url}
\usepackage{arydshln}

\usepackage{accents}
\usepackage{wrapfig}
\usepackage{algpseudocode}
\usepackage{algorithm}
\algdef{SE}[PROCEDURE]{Procedure}{EndProcedure}%
   [2]{\algorithmicprocedure\ \textproc{#1}\ifthenelse{\equal{#2}{}}{}{(#2)}}%
   {\algorithmicend\ \algorithmicprocedure}%
\algdef{SE}[FUNCTION]{Function}{EndFunction}%
   [2]{\algorithmicfunction\ \textproc{#1}\ifthenelse{\equal{#2}{}}{}{(#2)}}%
   {\algorithmicend\ \algorithmicfunction}%

%% file: macros.tex
\def\*#1{\boldsymbol{#1}}

\newtheorem{lem}{Lemma}[section]

\newcommand{\tw}[0]{\textwidth}
\newcommand{\igr}[2]{\includegraphics[clip,width=#1\tw]{#2}}
\newcommand{\cigr}[2]{\centering \includegraphics[clip,width=#1\tw]{#2}}

\newcommand{\pd}[2]{{\frac{\partial #1}{\partial #2}}}
\newcommand{\argmax}{\mathop{\text{argmax}}}

\newcommand{\eq}[1]{(\ref{#1})}

\newcommand{\lw}[1]{\smash{\lower2.ex\hbox{#1}}}

\newcommand{\rbr}[1]{\left(#1\right)}

\newcommand{\RR}{\mathbb{R}}

\newcommand{\EE}{\mathbb{E}}

\newcommand{\cD}{{\cal D}}

\newcommand{\cF}{{\cal F}}

\newcommand{\cJ}{{\cal J}}

\newcommand{\cM}{{\cal M}}
\newcommand{\cN}{{\cal N}}

\newcommand{\cQ}{{\cal Q}}

\newcommand{\cX}{{\cal X}}

\newcommand{\cZ}{{\cal Z}}